%% file: JAIR_Example_Template.tex
\documentclass[screen, nonacm, authorversion]{jair}

\RequirePackage[
  datamodel=acmdatamodel,
  style=acmauthoryear,
  backend=biber,
  giveninits=true,
  uniquename=init
  ]{biblatex}

\usepackage{graphicx}
\usepackage{microtype}
\usepackage{subfigure}
\usepackage{booktabs} 
\usepackage{subcaption}
\usepackage{tikz}
\usetikzlibrary{shapes, positioning, arrows}

\usepackage{amsmath}
\usepackage{mathtools}
\usepackage{amsthm}

\usepackage[capitalize,noabbrev]{cleveref}

\usepackage{xcolor}

\definecolor{newTextColor}{rgb}{0, 0, 0.75}

\definecolor{VeryNewTextColor}{rgb}{0.75, 0, 0}

\makeatletter
\def\@journalName{Preprint}
\def\@journalNameShort{Preprint}
\def\@JAIRAE{}
\def\@JAIRTrack{}
\makeatother

\usepackage{fancyhdr}

\makeatletter
\AtBeginDocument{%
  \fancypagestyle{mystyle}{%
    \fancyhf{}%
    \fancyfoot[C]{\thepage}
  }%
  \pagestyle{mystyle}
}
\makeatother

\begin{document}

\title[Can LLMs Replace Economic Choice Prediction Labs?]{Can LLMs Replace Economic Choice Prediction Labs? The Case of Language-based Persuasion Games}



\author{Eilam Shapira}
\authornote{Corresponding Author.}
\orcid{0009-0007-4536-5758}
\email{eilam.shapira@gmail.com}
\affiliation{%
  \institution{Technion - Israel Institute of Technology}
  \city{Haifa}
  \country{Israel}
}

\author{Omer Madmon}
\orcid{0009-0001-4009-0368}
\email{omermadmon@gmail.com}
\affiliation{%
  \institution{Technion - Israel Institute of Technology}
    \city{Haifa}
  \country{Israel}
}

\author{Roi Reichart}
\orcid{0000-0001-6918-0554}
\email{roireichart@gmail.com}
\affiliation{%
  \institution{Technion - Israel Institute of Technology}
    \city{Haifa}
  \country{Israel}
}

\author{Moshe Tennenholtz}
\orcid{0000-0002-9459-5388}
\email{moshe.tennenholtz@gmail.com}
\affiliation{%
  \institution{Technion - Israel Institute of Technology}
    \city{Haifa}
  \country{Israel}
}

\renewcommand{\shortauthors}{Shapira, Madmon, Reichart \& Tennenholtz}

\begin{abstract}
\input{abstract}
\end{abstract}



\received{20 February 2025}
\received[revised]{20 October 2025}

\maketitle

\thispagestyle{mystyle}

\input{paper}

\printbibliography
\input{appendix}

\end{document}

%% file: abstract.tex
Human choice prediction in economic contexts is crucial for applications in marketing, finance, public policy, and more. This task, however, is often constrained by the difficulties in acquiring human choice data.
With most experimental economics studies focusing on simple choice settings, the AI community has explored whether LLMs can substitute for humans in these predictions and examined more complex experimental economics settings. However, a key question remains: can LLMs generate training data for human choice prediction? We explore this in language-based persuasion games, a complex economic setting involving natural language in strategic interactions. Our experiments show that models trained on LLM-generated data can effectively predict human behavior in these games and even outperform models trained on actual human data. Beyond data generation, we investigate the dual role of LLMs as both data generators and predictors, introducing a comprehensive empirical study on the effectiveness of utilizing LLMs for data generation, human choice prediction, or both. We then utilize our choice prediction framework to analyze how strategic factors shape decision-making, showing that interaction history (rather than linguistic sentiment alone) plays a key role in predicting human decision-making in repeated interactions. Particularly, when LLMs capture history-dependent decision patterns similarly to humans, their predictive success improves substantially. Finally, we demonstrate the robustness of our findings across alternative persuasion-game settings, highlighting the broader potential of using LLM-generated data to model human decision-making.\footnote{Our data and code will be released upon acceptance.}

%% file: paper.tex
\section{Introduction}

In the digital economy, online platforms have become central to the interaction between consumers and service providers. These platforms, such as e-commerce websites, travel booking sites, and online marketplaces, facilitate a dynamic exchange where service providers present their offerings and consumers make purchasing decisions based on the provided information. A specific example of such an interaction is seen on Booking.com, a popular travel booking platform. On Booking.com, hotel owners (sellers) aim to persuade potential customers to choose their hotels by presenting information, often expressed in natural language, such as textual descriptions and reviews.

These interactions are often repeated, with sellers employing different strategies to attract and retain customers. For instance, some sellers might consistently highlight positive reviews to appeal to potential buyers, even when there are some downsides to their services, a practice can be seen as a \texttt{Greedy} persuasion strategy. Other sellers may adopt a more careful approach that aims to build long-term trust and maintain a solid reputation, by presenting both positive and negative aspects of the hotel, conditional on its true quality. This can be seen as adopting an \texttt{Honest} persuasion strategy.

A platform such as Booking.com is often interested in accurately anticipating consumers' behavior when facing different types of persuasion strategies employed by sellers. By accurately predicting behavior, the platform can asses the expected satisfaction and engagement of its users, and the impact of particular sellers on the overall users' welfare.

In the economic literature, the concept of persuasion has been extensively studied, particularly within the framework of \emph{persuasion games} \shortcite{aumann1995repeated,farrell1996cheap,kamenica2011bayesian}. These games involve strategic interactions where a \emph{sender}, possessing private information (the actual quality of the hotel, in our example), aims to influence the decision of a \emph{decision-maker} through selective information disclosure (in our case, selection of the review to be presented). Reputation indeed plays a significant role in repeated persuasion games, as demonstrated in previous work \shortcite{kim1996cheap,aumann2003long,best2022persuasion,arieli2023reputation}.

Traditional economic models, however, often abstract these interactions into simplified messages, lacking the nuance and complexity of natural language communication.\footnote{In economic modeling, the message typically influences the receiver's beliefs solely through the application of Bayes' rule. The content of the message itself is usually abstracted away, meaning that the specific language or framing of the message does not play a role in the analysis. For example, two messages may be called 'good' and 'bad' for ease of exposition, but if they were called $m_1$ and $m_2$ nothing would have changed in the analysis, as long as the sender follows the same information revelation strategy in providing them.} This abstraction limits the applicability of these models to real-world scenarios where language plays a critical role. Consequently, there is a growing need for interdisciplinary research to better understand and predict human decision-making in such contexts.

Indeed, the study of \emph{language-based persuasion games} has recently gained popularity within the natural language processing community. While some previous work focused on optimizing the sender's strategy \shortcite{raifer-etal-2022-designing}, another important, complementary line of research focused on predicting the behavior of human decision-makers against a given set of reasonable persuasion strategies \shortcite{apel2022predicting,shapira2025human}. Importantly, even without considering a particular business application (e.g., Booking.com), predicting human behavior in strategic interactions has a huge intellectual value for the field of behavioral economics, and its understanding will contribute to the overall understanding of human behavior.

In the language-based persuasion game considered by these works, a travel agent (expert) is trying to persuade a decision-maker (DM) towards accepting their hotel offer, by presenting the decision-maker with a textual review of the hotel, selected from a given available set of reviews. 
The true quality of the hotel is the expert's private information, and the DM benefits from accepting the deal only if the hotel is of high quality.\footnote{As explained in Section \ref{sec:task def}, the true quality of the hotel is determined by numerical scores associated with its textual reviews.
}
The game consists of several rounds played between the same (bot) expert and (human) DM pair, which means that a DM facing a specific expert strategy can potentially learn and adapt over time, based on past experience.
\citet{apel2022predicting} was the first to introduce the study of human choice prediction in the above game, and employed various machine learning (ML) techniques to solve it. \citet{shapira2025human} then studied off-policy evaluation in a similar setup, i.e., predicting human decisions when faced with an expert strategy that was not observed during training time.

Importantly, existing methods for human choice prediction in language-based persuasion games, as well as in other economic contexts \shortcite{plonsky2017psychological,rosenfeld2018predicting,plonsky2019predicting}, rely on training ML models on a human choice dataset. Unfortunately, the collection, storage and usage of human choice data are often fraught with various challenges: first, it requires the development of designated tools and environments (e.g., a mobile application with a user-friendly interface). Moreover, privacy and legal issues must be addressed to permit the collection, storage, and utilization of this data.\footnote{See \citet{acquisti2016economics} for a survey on the economics of privacy.} 
Human data collection is also a long and tedious process, frequently resulting in issues like participant inattention, which can compromise data quality and must be addressed.
These challenges often lead to a process that is extremely inefficient, expensive, and time-consuming.

Meanwhile, Large Language Models (LLMs) have made significant progress in recent years, demonstrating capabilities across a broad spectrum of applications, including text summarization, machine translation, sentiment analysis, and more \shortcite{brown2020language_lang1,zhang2023extractive_lang2,peng2023towards_lang3,susnjak2023applying_lang4,wang2023chatgpt_lang5,openai2023gpt4}. Moreover, recent study demonstrates how LLM-based agents can successfully function as decision-makers in economic and strategic environments, in which agents aim to maximize their gain from – in a complex, possible multiagent interaction \shortcite{xi2023rise}. In the context of human choice prediction, using LLM-based agents to generate synthetic but realistic data represents a groundbreaking proposition. If LLMs can effectively mimic human behavior in these economic settings, they could offer a cost-effective, efficient, and scalable alternative to traditional methods for training human choice prediction models. Importantly, in most real-life scenarios, the generation of a large LLM-based sample is significantly easier than obtaining even a small human choice dataset.

\subsection{Our Contribution}
In this paper, we introduce a novel in-depth study of the predictive power of LLM-generated data to human choice prediction in language-based persuasion games. Sections \ref{sec:task def}, \ref{sec:data collection}, and \ref{sec:models} provide details regarding the game setup and task definition, human and LLM data collection, and prediction models and baselines used in the experiments, respectively.

We show that a prediction model trained on a dataset generated by LLM-based players can accurately predict human choice behavior. In fact, it can even outperform a model trained on actual human choice data, for a large enough sample size. 
We also demonstrate how combining LLM-generated training data with actual human-generated data can further improve the accuracy of the prediction task (Section \ref{sec:main res}).
This success, however, comes at a cost: we note that those predictors who rely solely on LLM-generated data are significantly less calibrated compared to those trained on human data.
Nevertheless, we show that combining LLM-generated and human data leads to the most accurate and calibrated model (Section \ref{sec:calibration}).
We further demonstrate the robustness of our approach by evaluating its effectiveness against different expert strategies separately, and draw insightful conclusions regarding choice prediction against some particularly important strategies (Section \ref{sec:per expert main res}).

\begin{figure*}[t!]
  \centering \includegraphics[width=1\textwidth]{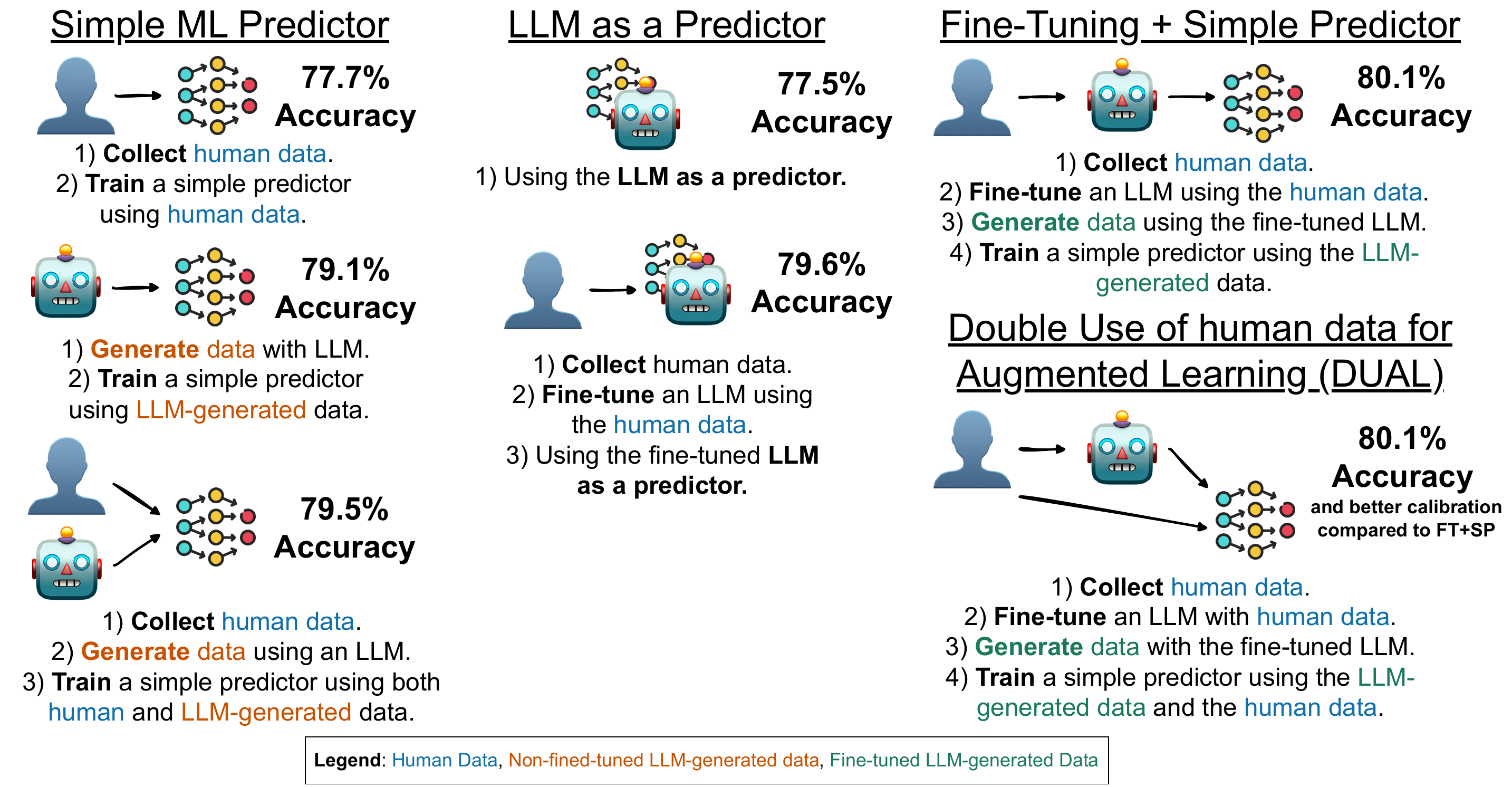}
  \caption{Results for the prediction task introduced in Section~\ref{sec:task def}, comparing alternative ways to use data from the 110 human players and from the LLM-generated players. Left: a simple ML predictor (see Section~\ref{sec:main res}) trained under three regimes—human-only data, LLM-generated data, and a mixed dataset. Middle: using the LLM itself as the predictor (see Section~\ref{sec:tuning}): an off-the-shelf LLM versus the same LLM fine-tuned on the human data. Right: fine-tuning an LLM and then using it to generate training data for a simple predictor (see Section~\ref{sec:tuning}). In the Double Use of human data for Augmented Learning (DUAL) variant, the human data used for fine-tuning is reused to train the simple ML predictor. While DUAL does not improve accuracy over Fine-Tuning, it nearly halves the expected calibration error.}
\label{fig:all_results}
\end{figure*}

We next explore alternative ways of integrating LLMs into the human choice prediction pipeline (Section \ref{sec:tuning}). While the main analysis in this paper focuses on using LLMs solely for training data generation, we emphasize that they can also serve directly as prediction models. We further investigate the potential of fine-tuning in these contexts, noting that the dual role LLMs can play (both as data generators and as predictors) opens up several strategies for applying fine-tuning. 
First, we show that fine-tuning the LLM-based data generator on human data improves the quality of the generated data, leading to improved prediction accuracy when predictors are trained on it.
Next, we demonstrate that the same human data used for fine-tuning can also be incorporated directly into the training set, yielding a mixed dataset of human and LLM-generated examples that boosts calibration while preserving the same predictive performance in terms of accuracy. We refer to this latter approach as \emph{Double Use of human data for Augmented Learning (DUAL)}. Figure \ref{fig:all_results} summarizes this spectrum of prediction strategies, and Section \ref{sec:tuning} compares and discusses their effectiveness in detail.

We then apply our choice prediction framework to study behavioral patterns that emerge in the persuasion game (Section \ref{sec:history-importance}). We find that decision-makers’ long-term behavior in this environment is primarily shaped by two distinct factors. The first is the \emph{natural-language signal} -- the current hotel review presented by the expert to describe the hotel. The second is the \emph{interaction history} with the same expert across repeated rounds of the game. This historical component captures the \emph{strategic} dimension of decision-making, reflecting behavioral dynamics such as trust, cooperation, and punishment. For example, decision-makers may interpret the same current signal differently depending on whether past positive reviews proved reliable or deceptive. Our analysis demonstrates that interaction history plays a crucial role in predicting and explaining human behavior.

In Section \ref{subsec:similarity-predictive-success} we show that whenever LLMs’ decision patterns resemble those of humans \emph{in how past interactions influence future choices}, the predictive performance of models trained on LLM-generated data improves. In contrast, when the LLM relies primarily on the linguistic signals (rather than on past interactions), the predictive value of its generated data decreases -- indicating that interpreting human choice as a naive sentiment analysis task is misleading, and capturing behavioral patterns in repeated interactions requires modeling choice history.
In Section \ref{subsec:markov-llm}, we experiment with varying the length of interaction history available to the LLM when making decisions. We find that when the LLM’s context is restricted to only one or two interaction turns, the data it generates leads to a noticeable drop in the predictive accuracy of the predictor trained on it, compared to cases where the full interaction history is provided.

To assess the generalizability of our findings, we also present preliminary experimental results using an alternative framework for language-based persuasion games proposed by \citet{shapira2024glee}. In this setting, the game is framed as a repeated interaction between a buyer and a seller, where the seller’s behavior is driven directly by an LLM rather than a predefined rule-based policy. 
That is, the framework of \citet{shapira2024glee} differs from ours in terms of both \emph{framing} (buyer-seller interactions vs. hotel adoption scenarios) and \emph{persuasion strategies} (LLM-based vs. rule-based).
Our experiments demonstrate that the potential of LLM-generated data for human choice prediction extends beyond the well-studied hotel experts game \cite{apel2022predicting,raifer-etal-2022-designing,shapira2025human}, further supporting the robustness and reliability of our approach (Section \ref{sec:general_persuasion}).

In Section \ref{sec:disc}, we conclude by discussing the potential limitations of our study, outlining promising directions for future research, and highlighting key ethical considerations that must be addressed when studying human choice prediction using AI-based tools.

\subsection{Related Work}
\label{sec:related}

\paragraph{LLMs and human behavior} Recent studies explore the ability of LLMs to replicate human behavior \cite{cui2025large}. Their integration into scientific workflows has sparked diverse perspectives and debates, as discussed in \citet{doi:10.1073/pnas.2401227121}.
Previous works demonstrate the abilities of LLMs to solve stumpers \shortcite{goldstein2023decoding}, take creativity tests \shortcite{stevenson2022putting}, and simulate human samples from sub-populations in social science research \shortcite{argyle2023out}. 
Another line of research focused on exploring whether and when LLMs can replace human participants in psychological science \shortcite{aher2023using,hussain2023tutorial,dillion2023can_p1,demszky2023using,taubenfeld2024systematicbiasesllmsimulations}.
\citet{amirizaniani2024llmsexhibithumanlikereasoning} evaluate the Theory of Mind capabilities of LLMs through open-ended responses.
Closer to our work, \citet{horton2023large_p2} evaluated LLMs in experiments motivated by classical behavioral economics experiments of \citet{kahneman1986fairness}, \citet{samuelson1988status}, and \citet{charness2002understanding}.
\citet{sreedhar2024simulatinghumanstrategicbehavior} used LLMs to simulate human strategic behavior in the classical ultimatum game.
Interestingly, \citet{macmillan2024ir} show that although LLMs, like humans, deviate from fully rational behavior, the nature of their deviations differs from human patterns, highlighting the need for caution when using LLMs to simulate human behavior.
These growing lines of research inspired us to ask whether the ability of LLMs to behave like humans implies they can function as training data generators for human choice prediction. Importantly, we demonstrate this novel approach in a well-motivated economic setup, where, in contrast to previous work, natural language plays a crucial role in the interaction.

\paragraph{LLMs as data generators} 
There is a vast literature on training ML models using synthetic data \shortcite{synth_1,synth_2,synth_3,synth_4,synth_5,liu2024best_synth6,long2024llms}. The recent rise of LLMs offers promising advancements in this area by providing scalable and high-quality methods for generating synthetic data. LLMs have been successfully used as data generators for tabular data \shortcite{borisov2022language}, medical dialogue summarization \shortcite{chintagunta2021medically}, text classification \shortcite{meng2022generating,ye2022zerogen}, and more. 
LLMs have also been used to annotate data \shortcite{chiang2023can_p3,thomas2023large_p4}, improve document ranking \shortcite{askari2023generating_p6}, and replace human judges in NLP tasks \shortcite{bavaresco2024llmsinsteadhumanjudges}. 
At the same time, training models on synthetic data can also decrease model performance in certain contexts, as shown by \citet{shumailov2023curse} and statistically analyzed by \citet{seddik2024bad_synth_bad}.
To the best of our knowledge, we are the first to demonstrate the effectiveness of LLMs as training data generators for human choice prediction.

\paragraph{LLMs as rational agents} LLMs have also emerged as potential rational agents in economic setups \cite{feng2024survey,sun2025game}. This new paradigm marks a significant shift from past approaches where algorithms devoid of language capabilities were utilized for solving complex games such as Chess and Go \shortcite{silver2017mastering,campbell2002deep}.
LLMs offer a novel perspective by acting as rational agents, as demonstrated in previous work: \citet{guo2024economics} show that LLMs may converge to Nash-equilibrium strategies, and \citet{akata2023playing} demonstrate how LLMs tend to cooperate in repeated games.
\citet{park2024llm} study the long-term behavior of LLMs through the lens of online learning and regret-minimization, and propose a custom unsupervised loss to promote no-regret behavior.
Beyond their role as standalone decision-makers, recent work by \citet{10.1145/3699824.3699832} suggests that the performance of LLMs can be significantly enhanced by modeling their interactions with human users in a game-theoretic framework.
There is a vast recent literature on concrete applications of LLMs as rational agents, including negotiation \shortcite{fu2023improving}, task-oriented dialogue handling \shortcite{ulmer2024bootstrapping}, response under strategic classification \shortcite{xie2025strategic}, 
competitive content generation \shortcite{mordo2025rlrf},
information design \shortcite{duetting2025information},
and more.
Here we leverage these capabilities of LLMs to simulate strategic behavior in a fundamental economic setup in which language is coupled with strategic behavior. We then use this simulated data to predict human behavior.

\section{Task Definition}
\label{sec:task def}

\paragraph{The language-based persuasion game} We begin by describing the language-based persuasion game presented by \citet{apel2022predicting} and \citet{shapira2025human}, which is used to define our human choice prediction task. The game consists of two parties, an expert (she) and a decision-maker (DM; he), interacting for $T$ rounds. At the beginning of each round, the expert is presented with $R$ pairs of textual reviews and numerical scores (in the 1-10 range), describing a hotel the expert aims to promote in the current round. Denote by $r^t_i$ and $s^t_i$ the $i$'th textual review and score presented at time $t$, respectively. 
The expert's hotel in round $t$ is considered of high quality if its average score is at least $\tau$, and otherwise is considered of low quality. We define hotel's quality to be $q_t = \mathbf{I} \bigg( \frac{1}{R} \sum_{i=1}^R s^t_i \ge \tau \bigg)$, where $\mathbf{I}(\cdot)$ is the indicator function. 

\begin{figure*}[h]
  \centering \includegraphics[width=1\textwidth]{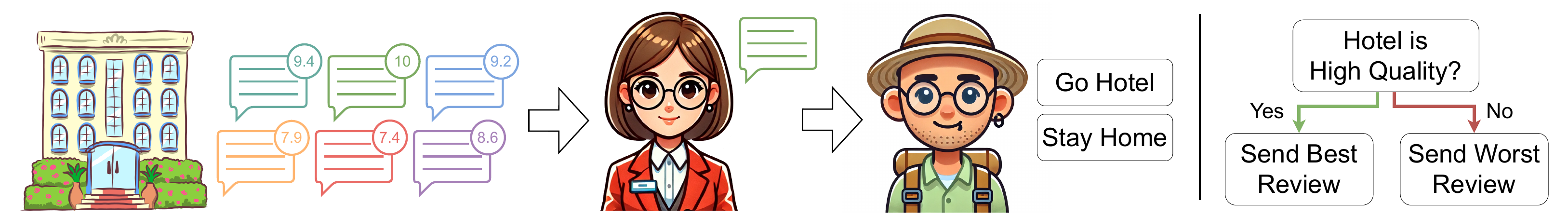}
  \caption{\textbf{Left:} Illustration of a single round in the language-based persuasion game. First, the expert observes the interaction history of previous rounds (does not appear in the illustration), as well as the current hotel's review-score pairs. She chooses a single review within this set according to her predefined strategy and sends it to the DM. Then, the DM observes this review (as well as the entire interaction history) and chooses an action. Lastly, both agents get their payoffs based on the DM's action and the hotel's true quality. \textbf{Right:} An example of an expert strategy.
  }
\label{fig:gametree}
\label{fig:binary_tree}
\end{figure*}

Then, the expert selects a single textual review $r_i$ and sends it to the DM. Note that the full set of review-score pairs is the expert's private information and is not available for the DM. Upon observing the message, the DM decides whether to go to the hotel ($a_t = 1$) or not ($a_t = 0$).
Lastly, the two players gain utility depending on the quality of the hotel $q_t$ and the DM's action $a_t$. The expert's utility is simply given by $u(a_t) = a_t$, meaning she always gains if the DM goes to the hotel, regardless of its actual quality. The DM, however, only benefits from opting in when the hotel is of high quality. His utility function is given by
$
    v(a_t, q_t) = \mathbf{I}(a_t = q_t)
$. Figure \ref{fig:gametree} (left) illustrates a single round in the game. The success criteria for both players is gaining as high cumulative utility as possible throughout the rounds of the game.

\paragraph{Expert strategies} 
We restrict our attention to the representative set of strategies considered first by \citet{shapira2025human}. These strategies are simple and intuitive, and can be represented as binary decision trees. Figure \ref{fig:binary_tree} (right) demonstrates such an example strategy of the expert (this is the \texttt{Honest} strategy discussed in the introduction). We highlight that all other strategies also have intuitive behavioral interpretations, and they differ from each other by the extent to which each strategy balances between building trust and exploiting trust.
We note that any expert strategy is either \emph{naive} (the rule according to which the presented review is selected is independent of both the hotel's true quality and the interaction history); \emph{stationary} (the selection rule depends solely on the hotel's quality, but not on the history); or \emph{adaptive} (the rule depends on the interaction history). Importantly, our representative set of strategies contains strategies belonging to all three groups.
For completeness, in Appendix \ref{app:experts} we formally introduce all strategies and discuss their motivations and behavioral interpretations.


\paragraph{The human choice prediction task} Given a dataset comprising multiple expert-DM interactions (i.e., multiple games, each consisting of multiple rounds), the task is to predict how other human DMs will engage in the game against the same experts that appear in the train set. 
Given an available training dataset of expert-DM interactions in language-based persuasion games, the goal is to predict the behavior of human DMs against the same set of expert strategies that generated the training data (this stands in contrast to \citet{shapira2025human}, in which there is a mismatch between training-time and test-time experts). However, unlike previous work on prediction in persuasion games, we aim to evaluate the prediction quality when the training data does not consist of any actual human DMs data, and instead consists of data generated by LLM-based players.
Similarly to \citet{shapira2025human}, we evaluate a prediction model with respect to the per-DM per-expert average accuracy.

\section{Data Collection}
\label{sec:data collection}



\paragraph{Human dataset} We use the human-bot interaction dataset of \citet{shapira2025human}.\footnote{http://github.com/eilamshapira/HumanChoicePrediction} This dataset was collected via a mobile application (published on both Apple's App Store and Google Play), in which human DMs interact with 6 experts, where a \emph{stage} refers to an interaction with one of these experts. In each stage, the human DM plays \emph{multiple games} against the same bot expert, denoted $e_i$ (recall that each game has $T$ \emph{rounds}, where they set $T=10$). The $i$'th stage ends whenever the human player reaches a pre-defined cumulative utility $\bar{v}_i$ in a $T-$round single game.\footnote{For our prediction task, we consider only the first two games each human player played against each expert (or a single game if the player completed the stage in a single game). 
We restrict our analysis to the initial games due to a noticeable decline in participants' response times in subsequent games, suggesting reduced attention and non-strategic behavior. 
This approach is widely accepted in social sciences and experimental economics, see e.g. \citet{Rubinstein_2013}.}
Human data was collected from May 2022 to November 2022. The human dataset contains 71,579 decisions made by 210 distinct human DMs that completed the game, i.e. completed all six stages.\footnote{Players were incentivized to complete all six stages in various ways, including participation in lotteries and receiving academic credit for completing the game.}

\begin{figure}[h]
  \centering \includegraphics[width=260pt]{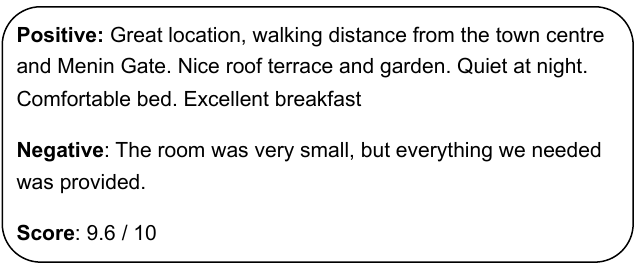}
  \caption{A sample review from the hotel reviews dataset.}
\label{fig:a_review}
\end{figure}

Hotel reviews were taken from Booking.com. The hotels and the game parameters were chosen such that each hotel has $R=7$ reviews, and about half of the hotels are defined to be of high quality (i.e., have an average score of at least $\tau = 8$). The hotel dataset used by \citet{shapira2025human} contained 1068 distinct hotels. Figure \ref{fig:a_review} provides an example review from the hotel reviews dataset, taken from \citet{shapira2025human}.

\paragraph{LLM datasets} 
To solve the human choice prediction task without using human-generated data in the train set, we created an LLM-generated dataset by replicating the human dataset data collection process, and replacing human DMs with LLMs. To do so, we implemented an identical pipeline with the exact same set of experts, hotels, and game parameters ($T$, $R$ and $\tau$).
We utilized 5 different state-of-the-art LLMs to generate the LLM datasets: Google's Chat-Bison \shortcite{anil2023palm} and Gemini-1.5 \shortcite{gemini_team_gemini_2024}, Alibaba's Qwen-2 72B \shortcite{yang_qwen2_2024}, and Meta's Llama-3 70B and 8B \shortcite{bhatt2024cyberseceval2widerangingcybersecurity}, all in their chat versions.
Similarly to the human choice dataset, each LLM-based player played two games against each expert (unless the LLM player completed the stage in the first game). We repeat this process over and over to create many LLM players. The prompt given to the LLMs is similar to the instructions and messages presented to the human players in the mobile application developed by \citet{shapira2025human}. Appendix \ref{app:conversasion} contains this prompt, as well as a conversation example. 


\paragraph{Persona diversification}
To diversify the LLM-generated dataset, we have associated each LLM player with a \emph{persona}. Each persona type specifies a typical behavior the LLM is instructed to follow. This is implemented by simply concatenating a sentence to the beginning of the initial prompt of the LLM player, that specifies the required behavior. For example, for the \texttt{optimistic} persona type, the initial prompt contains the game instructions (as they were written for the human players), followed by the following instruction: \emph{``behave like an optimistic person``}. 

\begin{table}[h]
    \caption{Persona types and the initial prompts of the LLM players.}
    \input{tables/llm_dataset}
    \label{tab:llms_dataset}
\end{table}

Table \ref{tab:llms_dataset} provides all descriptions and prompts for all 8 persona types we used. The first two persona types are the \texttt{optimistic} and \texttt{pessimistic} personas. The other six persona types were selected based on textual features extracted from Booking.com reviews, as described by \citet{apel2022predicting}. These reviews were represented using binary features that indicated whether various hotel aspects were discussed positively or negatively. We used these features to create different personas, each focusing on a specific aspect of the hotel.
For each language model, we generated decision data using all different persona types (personas, hereafter). For each specific persona, we generated a decisions dataset using a varying number of LLM-based players, as specified in Table \ref{tab:dataset_statistics}.

\input{tables/datasets_statistics}

\paragraph{Sample complexity} While using similar techniques has already been shown to encourage LLM agents to take a variety of social behaviors \shortcite{park2023generative}, in our context, we observed that persona diversification contributes to reducing the sample size required to obtain accurate predictions. Figure \ref{fig:personas_effect} shows the number of players required to achieve certain levels of accuracy in two different settings (here the LLM used to simulate DMs is Chat-Bison, and the prediction model is LSTM):
\begin{enumerate}
    \item \textbf{Without persona diversification.} LLM players were not assigned any persona type, i.e., no persona-associated prefix was added to their prompts.
    \item \textbf{With persona diversification.} Any LLM player was randomly assigned one of the 8 persona types defined above, and the corresponding prefix was added to its prompt.
\end{enumerate}
Notably, the persona diversification technique indeed improves sample complexity. Moreover, the gap between the two settings increases as the desired level of accuracy increases. After observing the phenomenon in Chat-Bison, which was the first LLM we considered, we decided to continue the data collection process using persona diversification, in order to reduce time and costs.

\input{figures/personas_effect_code}



\section {Models and Baseline}
\label{sec:models}

The primary goal of this work is to demonstrate the effectiveness of training a human choice prediction model using LLM-generated data. To do so, we compare the performance of prediction models trained on actual human-choice data with the same model trained on LLM-generated data, across various prediction model architectures. 
In addition, we compare to a baseline method in which data is collected using agents that rely solely on the sentiment analysis abilities of an LLM on the review text rather than a combination of both linguistic and behavioral understanding.
All prediction models are evaluated on the choice data corresponding to 100 human players in the dataset, randomly selected for 50 different test sets. Prediction models that used human training data were trained with $ K \in [32, 64, 110]$ human players which do not belong to the evaluated test set.

We consider four types of prediction models: LSTM \shortcite{6795963} and Mamba \shortcite{gu2024mambalineartimesequencemodeling} as sequential models, transformer \shortcite{vaswani_attention_2017} as an attention-based model, and XGBoost \shortcite{DBLP:journals/corr/ChenG16} as a history independent model.\footnote{We evaluate our methods with LSTM, Mamba and a 4-headed transformer, all with a learning rate of $4 \cdot 10^{-4}$, 64 hidden dimensions, and two layers. For XGBoost, we used 300 estimators with a max depth of 3.}
 All predictors use the same representation of the choice data and are being trained on data generated by the three different paradigms (human, LLM, and baseline). That is, throughout the experiments, we fix the prediction model architecture and only modify the data it is trained on. This enables the evaluation of the generated data quality in terms of enhancing the human choice prediction.

\paragraph{Models' input} Following \citet{apel2022predicting, shapira2025human}, each interaction round is represented using binary \textit{Engineered Features (EFs)} that capture topical, structural, and stylistic aspects of the review, as well as features describing the strategic context of the decision. 
We adopt EFs following the findings of \citet{shapira2025human}, where this representation proved more effective than BERT and GPT-4 text embeddings for predicting decision outcomes in our setup. The full list of features appears in Appendix D of \citet{shapira2025human}. 


\paragraph{Sentiment baseline} \label{sec:lig_baseline}
The baseline method, 
 instead of considering both the interaction history and the text, considers only the sentiment of the text.
 To implement this baseline, we first asked an LLM to predict the scores of all reviews in the dataset, and for each such review, we extracted the score distribution induced by the LLM using its logits (see Appendix \ref{app:review2score} for more details).
Then, we use this review-score joint distribution to simulate a choice dataset for the human choice prediction task. For every given review, we sample a score and apply the following decision rule: going to the hotel if and only if the sampled score is above the threshold $\tau$ that defines the hotel's quality. 

Importantly, unlike the LLM-based player, the baseline DM is forced to act solely based on the current linguistic signal (i.e., the agent's decision rule is history-independent). For instance, the baseline player cannot use punishment strategies in the form of "trust the expert until she turns out to have sent a great recommendation for a bad hotel". In contrast, both human and LLM players may condition their current actions not only on the current message but also on the interaction history, and potentially learn such complex strategies that involve patterns of cooperation and punishment.

For the baseline method, we use the same LLM that generated the data which achieved the best performance when looked both at the text and at the history. This baseline allows us to examine the advantage of a model trained on LLM data created with both linguistic and behavioral knowledge, compared to a model trained on an equal amount of data generated solely based on linguistic knowledge, thereby measuring the improvement in prediction that comes as a result of the LLM's economic understanding.

\section{Effectiveness of LLM-Generated Data}
\label{sec:main res}

\input{figures/main_result_code}

Figure \ref{fig:main_result} presents the accuracy obtained by the Qwen-2-72B LLM, which achieved the best performance for the prediction task.\footnote{
Note the atomic unit of evaluation is a \emph{decision} nor a \emph{player}. We report the number of players since each player makes a different number of decisions.}
It shows the results for varying training sizes, along with the sentiment analysis baseline.
The x-axis represents the number of different players used to create the expert-LLM players interaction dataset (logarithmic scale), while the y-axis displays the accuracy averaged over 50 runs. The results are presented with bootstrap confidence intervals, at a confidence level of 95\%, and are distinguished by the players in the train and test sets and the initial weights of the neural networks.
The grey horizontal lines indicate the accuracy achieved by a model trained with human players. The number of players used during training is displayed to the right of each line.

Notably, for all types of prediction models, LLM-generated data outperforms human data in the human choice task for a large enough sample size. Importantly, in most real-life applications, obtaining human data of sufficient sample size is significantly more complex and expensive compared to LLM data generation.

Table \ref{tab:offline_simulation} shows the accuracy obtained by prediction models trained on different data sources.
Aside from Qwen-2-72B, most LLMs (except Llama-3-8B) also generated data that allowed the training of a predictive model with quality surpassing that of a model trained with data from 16 humans.

\input{tables/main_results}

Next, we compare the performance of training with LLM-generated data to training with the sentiment baseline method. We recall that data generated by the baseline method is history-independent, in the sense that decisions of the current round are made based only on the current review score prediction. We note that using LLMs to simulate an end-to-end interaction (and, in particular, allowing such history-depending behavior) leads to better predictions of human choice behavior compared to this baseline.\footnote{The fact that this naive language-only baseline obtains decent results in terms of prediction accuracy is consistent with previous literature on sentiment analysis in economics environments, e.g. \shortcite{7955659,venkit2023sentiment}.} This implies that simulation interaction based only on linguistic aspects is insufficient for human choice prediction. In contrast, we hypothesize that our LLM-generated data encodes not only a linguistic understanding of the textual reviews but also an economic understanding of the interaction: LLM players may condition their current choice not only on the linguistic signal but also on the outcome of previous interactions, leading to a variety of behavioral patterns. Evidently, these patterns increase the predictive power compared to the baseline method. 

Since LSTM outperforms the other prediction models for each training data configuration, and Qwen-2-72B is the best data generator for our task, the results of all following experiments are shown only for the LSTM prediction model trained with the dataset created using the Qwen-2-72B LLM. 

\paragraph{Combination of LLM with human data}
\label{sec:human_and_llm}

We now investigate how mixing LLM and human data impacts performance.
Figure \ref{fig:human_and_llm} (left) shows the performance of different prediction models whose training set consisted of 110 human players, supplemented by a varying number of LLM players. It can be seen that enriching the human data with synthetic data significantly improved performance. Figure \ref{fig:human_and_llm} (right) shows the performance of a model trained with a varying number of human players supplemented by players generated by Qwen-2-72B, the best-performing LLM. It can be observed that a model trained on a mixture of LLM-based and human players outperforms a model trained on human data only. Hence, even if human data has already been collected, enriching it with LLM-based players will yield greater benefits than collecting additional data from humans.

\input{figures/human_and_llm_code}

\subsection{Calibration Error Analysis} 
\label{sec:calibration}

Our positive results throughout this section imply that human behavior can be predicted well without relying on actual human decision data during train time. However, it is crucial to recognize that there is no free lunch; the LLM-based approach falls short in calibration when compared to training on human data. Nevertheless, combining LLM-generated and human data results in the most calibrated model. Calibration is essential, for instance, when the output of one prediction task serves as input features for another. Trustworthy confidence levels from well-calibrated models prevent cascading errors.

\begin{figure}[h]
  \centering \includegraphics[width=0.65\linewidth]{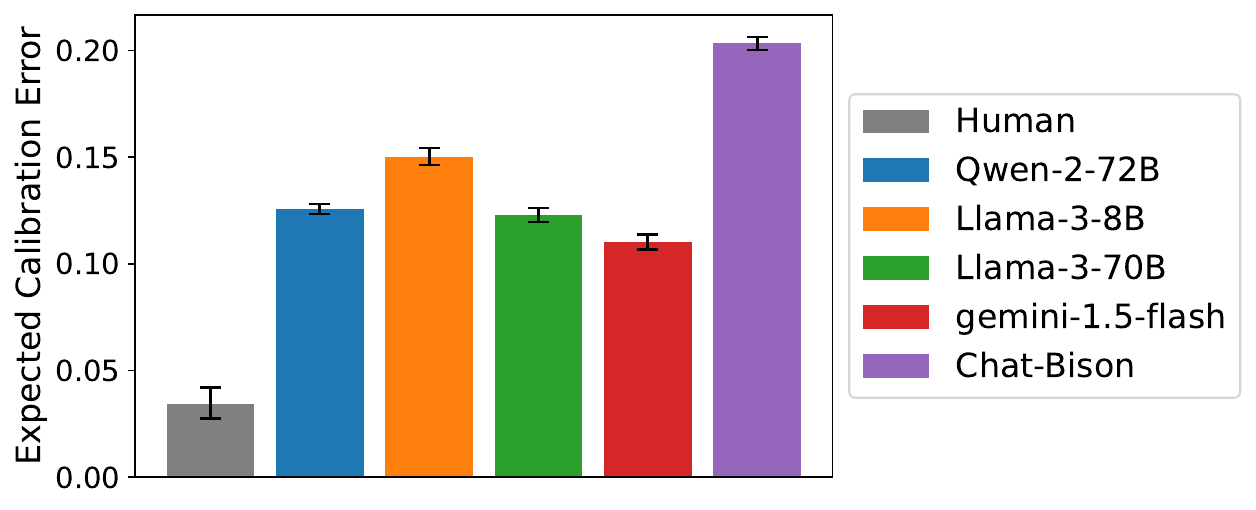}
  \caption{Expected Calibration Error of models trained by different datasets (lower is better). LLM models (colored bars) are trained on 1024 LLM players. The human model (gray bar) is trained on 110 players.
  Predictors trained on human data are better calibrated than any predictor trained on LLM-generated data.
  }
\label{fig:calibration}
\end{figure}

Expected Calibration Error (ECE) measures the difference between predicted probabilities and actual outcomes, ensuring that a model's confidence levels are accurate \shortcite{pakdaman_naeini_obtaining_2015}. 
Figure \ref{fig:calibration} shows the calibration obtained by prediction models trained on different data sources. Notably, the ECE values of models trained using LLM data are higher than those of models trained on human data, indicating a degradation in calibration in the prediction model trained on LLM data compared to a predictor trained on human data. Interestingly, we observe that the more successful the LLM used to generate the data in the prediction task, the more calibrated the prediction model becomes. 

\begin{figure}[h]
  \centering \includegraphics[width=\linewidth]{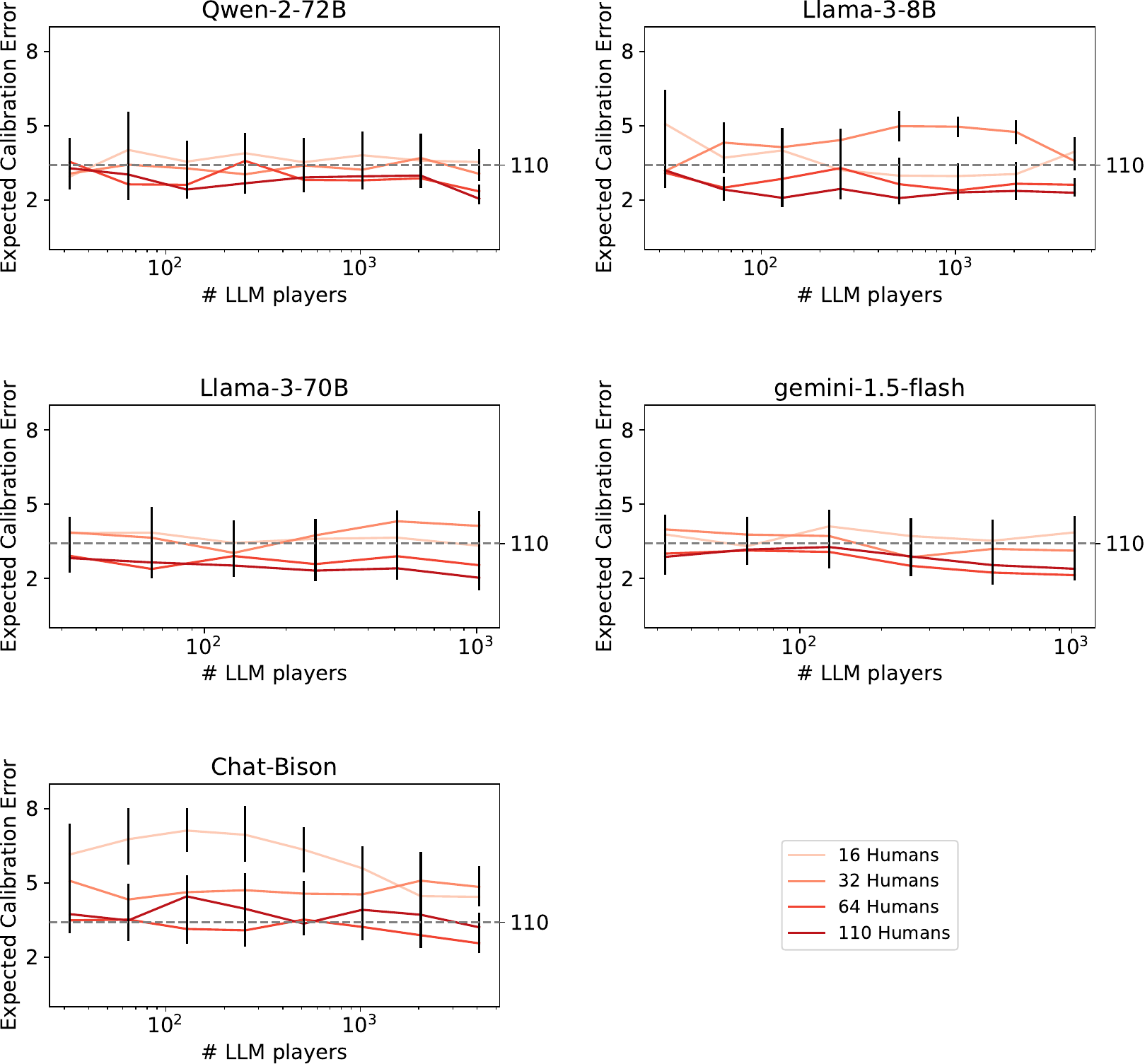}
  \caption{Expected Calibration Error of a model trained with different datasets (subplots) with different numbers of humans (color) and a varying number of LLM-players (x-axis). Gray lines: Models trained with data comprised of only human players. Mixing LLM-based and human data (darker lines) provides the best calibration error.}
\label{fig:calibration_with_human}
\end{figure}

Figure \ref{fig:calibration_with_human} shows the calibration obtained by prediction models trained on a combination of human data and data generated by different types of LLM players. As can be seen from the figure, augmenting human training data with LLM-generated data results in improved calibration compared to training on human data only.

Calibration is essential, for instance, when the output of one prediction task serves as input features for another. Trustworthy confidence levels from well-calibrated models prevent cascading errors. Awareness of our approach's calibration shortcomings encourages using re-calibration methods to improve model reliability. For instance, in a hypothetical scenario in which a system designer has access to a limited number of human decision data, she can decide whether to use this data in training time or allocate it to re-calibrate the model, depending on her accuracy/calibration requirements.

The calibration gap we observe between models trained on LLM-generated data and those trained on human-collected data aligns with findings in the machine learning literature. Models trained on data originating from a different distribution can sometimes achieve higher predictive accuracy but suffer from poorer calibration, as their confidence estimates no longer match empirical outcome frequencies. \citet{ovadia2019can} demonstrated that models well-calibrated on in-distribution data often become miscalibrated under distribution shift.

In our case, models trained on LLM-generated data achieve higher predictive accuracy primarily because a substantially larger volume of training data is available, which allows them to capture general behavioral patterns more effectively. However, these data reflect systematic but artificial regularities that do not perfectly match real human behavior, resulting in reduced confidence reliability. This trade-off underscores the importance of combining synthetic and human data to leverage the abundance of LLM-generated samples while preserving calibration with human-like behavior.

\subsection{Predicting Against a Specific Strategy}
\label{sec:per expert main res}

We have shown that our LLM-based data generation paradigm obtains high prediction accuracy, and even outperforms models trained on human data whenever sufficient synthetic data points are available. 
This section provides a more careful analysis of the effectiveness of this approach, by evaluating accuracy with respect to each \emph{individual} expert strategy separately. 
In Figure \ref{fig:strategies_performance}, each subplot corresponds to an individual expert strategy.\footnote{One can ask whether training a prediction model only on the individual expert interaction data (instead of using data that includes interactions of players with other experts) yields a higher accuracy. At the end of this section, we show that the answer is negative.}


\paragraph{Comparison to human-generated data} The LLM-based approach with Qwen-2-72B generated data always outperforms the models trained on 32 human players. Moreover, in some cases, it even outperforms models trained on 110 human players. These results indicate that the LLM-based approach is quite robust: not only that averaging over strategies yields a high accuracy in the prediction task, but also human actions against each strategy separately can be accurately predicted. 

\input{figures/two_strategies_performance_code}

\begin{figure*}[t]
  \centering \includegraphics[width=\linewidth]{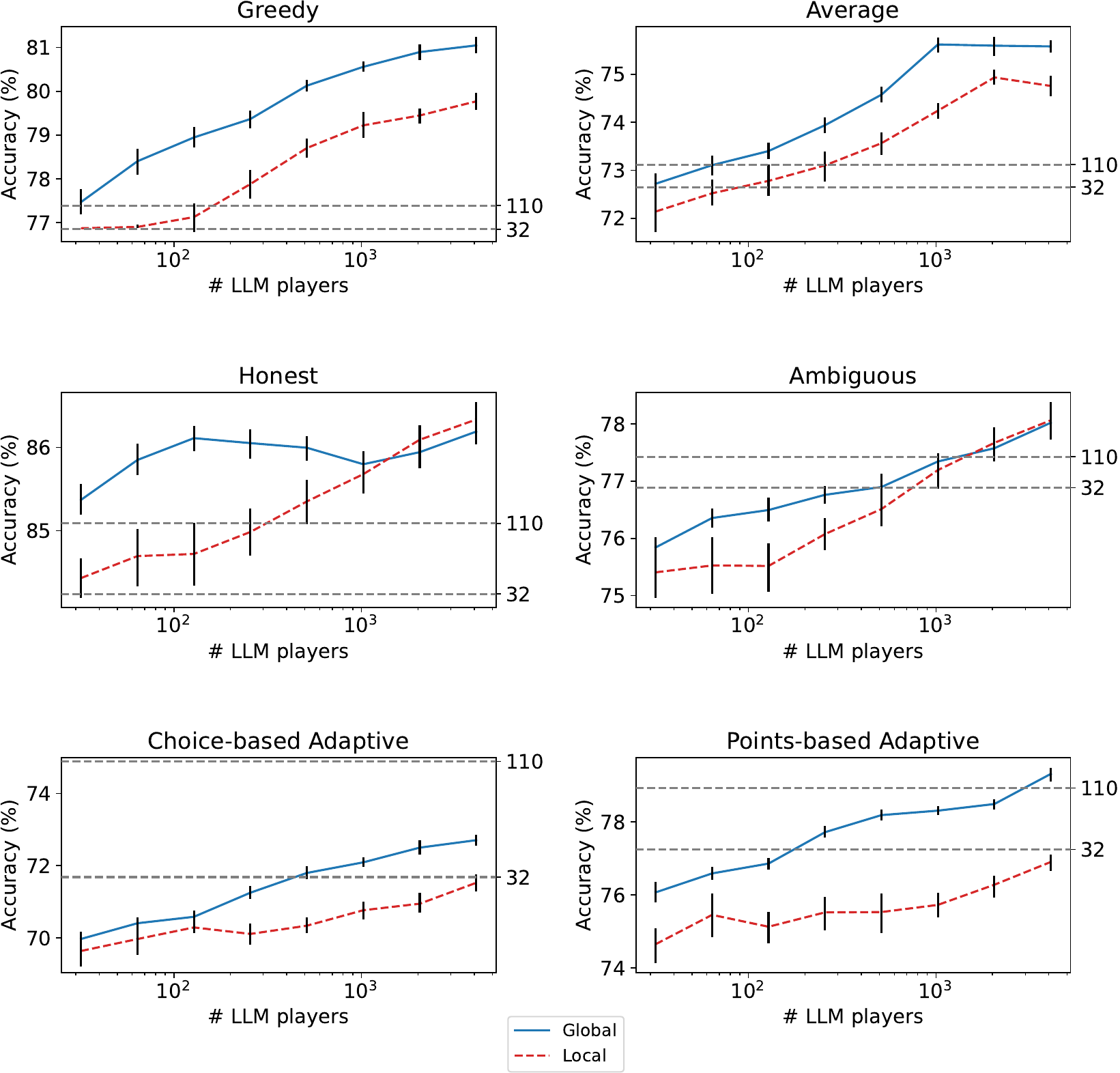}
  \caption{Accuracy on individual expert interactions only, obtained by prediction models trained on LLM-generated data (both local and global). For any individual expert, the global model outperforms the local model.}
\label{fig:all experts main res}
\end{figure*}

Interestingly, one of those cases in which our result is significantly effective, is the case of the \texttt{Greedy} strategy, which simply means the expert always sends the review corresponding to the highest score, regardless of the actual quality of the hotel, or the history of interactions with the DM player. This particular expert strategy is important for two reasons: first, it represents a very common and typical behavior of a naive expert who aims to greedily persuade non-sophisticated users towards opting in.
Second, it turns out that the \texttt{Greedy} expert strategy is very effective in terms of expert utility maximization.  \citet{raifer-etal-2022-designing} empirically studied a similar language-based persuasion game, and demonstrated the effectiveness of the \texttt{Greedy} strategy when used against human DMs. This aligns with our experimental setting, in which the \texttt{Greedy} expert strategy is shown to be the best expert strategy among all six strategies considered, against both human and LLM-based DMs (see Appendix \ref{app:per-exp}).
These arguments suggest that \texttt{Greedy} experts are expected to exist in many realistic cases. 

\paragraph{Comparison to the sentiment baseline} The LLM-based approach indeed outperforms the sentiment baseline in the majority of the cases. The only exception is the \texttt{Honest} strategy, on which the naive baseline outperforms both the LLM-based approach and the standard approach of training the predictor on human data. This is exactly the strategy discussed in the introduction and described in Figure \ref{fig:binary_tree} (Right), according to which the expert reveals the most positive review when the hotel is of high quality, and the most negative review when it is of low quality. In particular, this is a \emph{stationary} strategy (i.e., independent of the interaction history). Hence, from the DM perspective, once trust in the expert is established, the task of making a decision boils down to a simple sentiment analysis task, which is exactly what the sentiment baseline does. It is therefore reasonable that the baseline method performs well against this particular strategy, while against all other (non-truthful) strategies it is outperformed by LLM-based training.

\paragraph{Global vs. local per-expert prediction models}
In this section, we compared LLM-based to human-based training with respect to individual experts.
In this appendix, we answer the following natural question: whenever there is only a single expert for which human choice prediction is required, is it better to train a \emph{global model} (i.e., use all experts' interaction data to train the model) or a \emph{local model} (i.e., train only using interaction data corresponding to the individual expert)? This is a question of data quality vs. quantity trade-off: while a global model relies on more observations (many of which are somewhat irrelevant), the local model uses fewer observations, but all of them are collected with respect to the individual expert. Note that throughout this section we have always trained a global model (namely, included interaction with experts other than the individual expert within the training set), and the results of this experiment will justify this approach.

For each individual expert, Figure \ref{fig:all experts main res} shows both the results of a local and a global model, for models trained on Qwen-2-72B players. Evidently, the global model outperforms the local model for any possible individual expert. This implies that although the interaction data across expert strategies may be different, it is still beneficial in terms of enhancing the capabilities of an expert-specific prediction model.

\section{Advanced LLM Integration: Fine-Tuning for Prediction and Data Generation}\label{sec:tuning}

In the previous sections, we showed that LLMs can generate data of sufficiently high quality for predicting human decisions. The finding raises a crucial question: how can we best leverage a limited amount of authentic human data to further enhance predictive performance? While Section 5 showed the benefits of simply mixing datasets, this section investigates a more powerful approach: using the human data to refine the data-generator LLM itself through fine-tuning. The goal is to determine if specializing an LLM on human behavioral data can unlock a new level of predictive accuracy and address limitations such as the calibration gap identified earlier. To this end, we explore two primary strategies: (1) using the fine-tuned LLM directly as a choice predictor, and (2) employing it as an improved data generator. This investigation contributes to our broader research agenda by moving beyond simple data generation to more advanced hybrid methods, showcasing a path to maximizing predictive power.

\begin{figure}[t]
  \centering \includegraphics[width=0.65\linewidth]{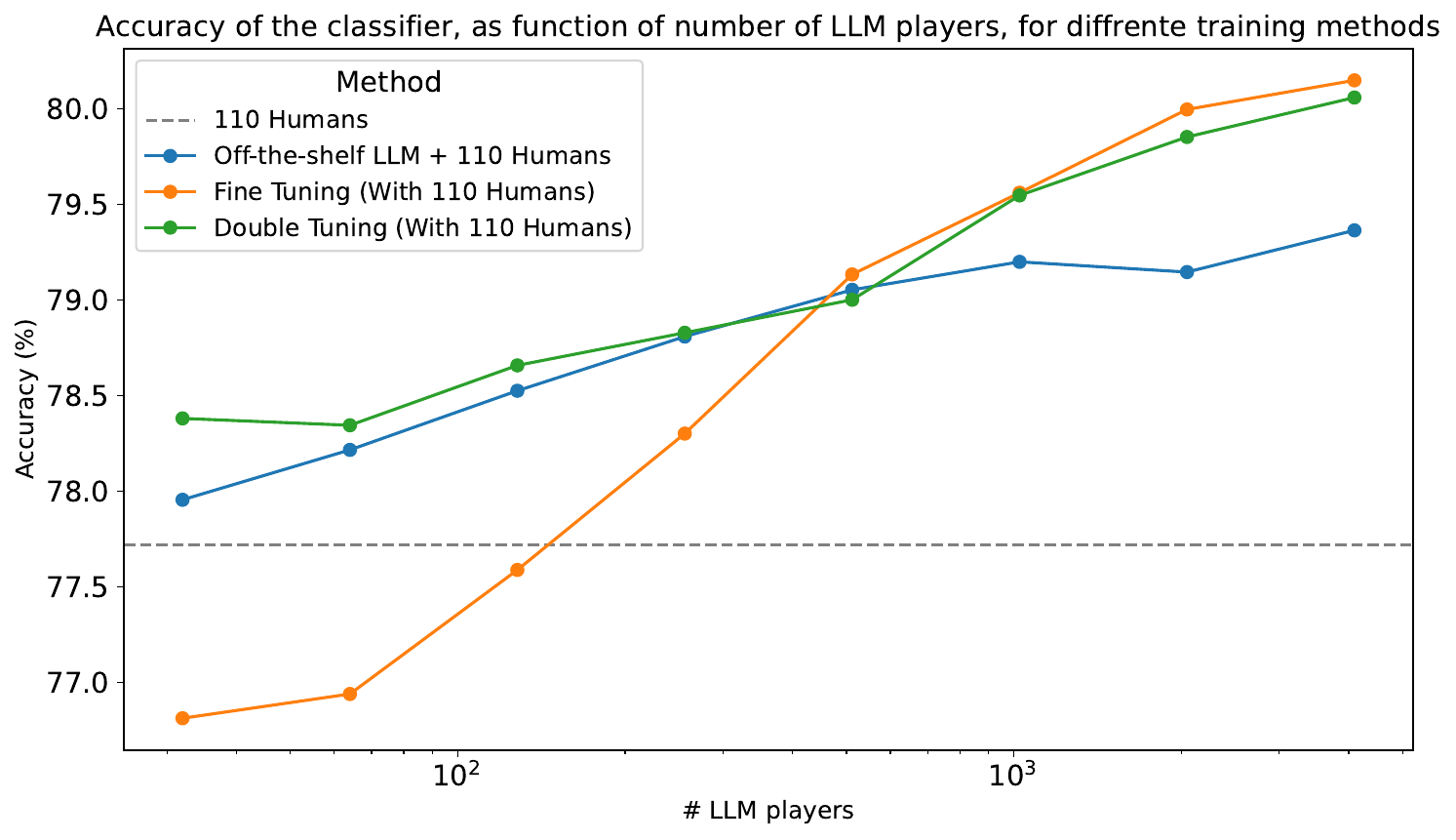}
  \caption{LSTM accuracy as a function of the number of LLM-generated players used for training, comparing three ways to combine LLM- and human-derived data: (i) \emph{Off-the-shelf LLM + 110 Humans}: generating players with an unmodified LLM, augmenting with data of 110 human players; (ii) \emph{Fine-Tuning}: fine-tuning an LLM on the 110 humans, then generating players; and (iii) \emph{Double Use of human data for Augmented Learning (DUAL)}: augment the Fine-Tuning data with the same dataset of 110 human players. 
  }
\label{fig:ft-generator}
\end{figure}

\paragraph{Fine-tuned LLM as a predictor}\label{par:ft-predictor}
One natural way of using an LLM for training predictive models of human decisions is to treat it as a predictor. After fine-tuning the model on a training dataset of human interactions, it is used for inference. During this inference phase, the LLM receives the full interaction history in the prompt, and its generated text is mapped to the forecast of the player’s next action.

We fine-tune Llama-3-8B on the training set comprising 110 human players. Despite Llama-3-8B producing the lowest-quality synthetic data among our evaluated LLMs prior to fine-tuning, fine-tuning on human data substantially improves predictive quality. While the non-fine-tuned model, when used as a zero-shot classifier, attains \(74.7\%\) action-prediction accuracy, the fine-tuned variant reaches \(79.6\%\). These results exceed the accuracy of every predictor trained on data from off-the-shelf LLMs, human data alone, or their combination. While this result confirms the known effectiveness of supervised fine-tuning, its significance lies in what it suggests about the underlying mechanism. 
This approach surpasses not only predictors trained on the same human data but also those augmented with synthetic data from more advanced, non-fine-tuned LLMs, highlighting a qualitative advantage in learning directly from human examples.

\paragraph{Fine-tuned LLM as a data generator}\label{subsec:ft-generator}
We next use the fine-tuned model as a data generator. Figure~\ref{fig:ft-generator} presents the performance of an LSTM trained on players generated by the fine-tuned Llama-3-8B (FT-Llama-3-8B) as a function of the number of generated players included in training. We report two variants:
\begin{itemize}
    \item \textbf{Fine-tuning}: the LSTM is trained solely on FT-Llama-3-8B-generated data.
    \item \textbf{Double Use of human data for Augmented Learning (DUAL)}: the LSTM is trained on a combination of FT-Llama-3-8B-generated data and the same 110 human players that were used to fine-tune the Llama-3-8B model.
\end{itemize}
Both variants are compared against (i) an LSTM trained on the 110 human players only and (ii) an LSTM trained on players generated by the strongest off-the-shelf LLM (Qwen-2-72B) plus the 110 human players. Fine-tuning yields a substantial improvement in predictive performance: the predictor trained solely on FT-Llama-3-8B-generated data achieves 80.1\% accuracy. While reusing the human-player data in DUAL does not further increase accuracy over fine-tuning and also achieves 80.1\% accuracy, it nearly halves calibration error (ECE), from \(0.15\) to \(0.08\).

These results offer several key insights into generating synthetic behavioral data. First, fine-tuning a large language model on even a modest amount of domain-specific human data can transform it into a highly effective data generator. The resulting synthetic data is of such high quality that a predictor trained exclusively on it surpasses models trained on the original human data or data from much larger, off-the-shelf LLMs. This demonstrates a powerful and efficient method for data augmentation.

Second, the \textbf{DUAL} approach, where the original human data is combined with the synthetic data, reveals a crucial advantage: while it does not improve raw predictive accuracy, it substantially improves the model's calibration. This suggests that the synthetic data effectively captures the predictive patterns for high accuracy, while reintroducing the original human data helps to ground the model's confidence, making its predictions more reliable. This two-stage process, fine-tuning an LLM for data generation and then training a downstream model on a mix of synthetic and real data, represents a promising strategy for building robust and well-calibrated predictive models in data-scarce environments.

\section{The Importance of History in Human Choice Prediction}\label{sec:history-importance}

In this section, we leverage our prediction framework to analyze the key drivers of human decision-making in the game. A central question in modeling human behavior is determining the relative importance of different informational cues. We specifically investigate whether a player's choices are better predicted by their historical pattern of actions or by the immediate content of the interaction, as conveyed through the linguistic signal of the review. By dissecting the predictive power of these two dimensions - behavioral history versus interaction content - we aim not only to report the outcomes of our experiments but also to provide deeper insights into what makes behavior human-like and how to best capture it in synthetic agents.

\subsection{Action Similarity and Predictive Success}\label{subsec:similarity-predictive-success}

We aim to understand which aspects of player behavior are most informative for generating synthetic data that improves predictive performance. Specifically, we develop a succinct representation of the decisions made by both human and LLM players in the games and compute the similarity between these representations. By comparing similarities across different behavioral groupings, we can infer which behavioral dimensions (e.g., historical context versus sentiment) better capture the underlying decision patterns that drive human-like play.

To quantify similarity we follow \citet{shapira2025human} and  compute the cosine similarity between vectors that encode the proportion of times a player chose to go to the hotel, aggregated under different grouping schemes:

\begin{enumerate}
    \item \textbf{Sentiment-based similarity}: Sentiment is represented by the numeric review score assigned on Booking.com. Reviews are bucketed into five groups, where $S$ denotes the sentiment score: (1) $S \ge 9.2$; (2) $S \in [8.3, 9.2)$; (3) $S \in (7.7, 8.3)$; (4) $S \in (6.8, 7.7]$; and (5) $S \le 6.8$.
    \item \textbf{History-based similarity}: History is represented by the Receiver’s two most recent actions and their outcomes in the two rounds preceding the current round.\footnote{The full list of 21 possible states appears in Appendix~\ref{app:history_options}.}
\end{enumerate}

\begin{figure}[h]
  \centering \includegraphics[width=\linewidth]{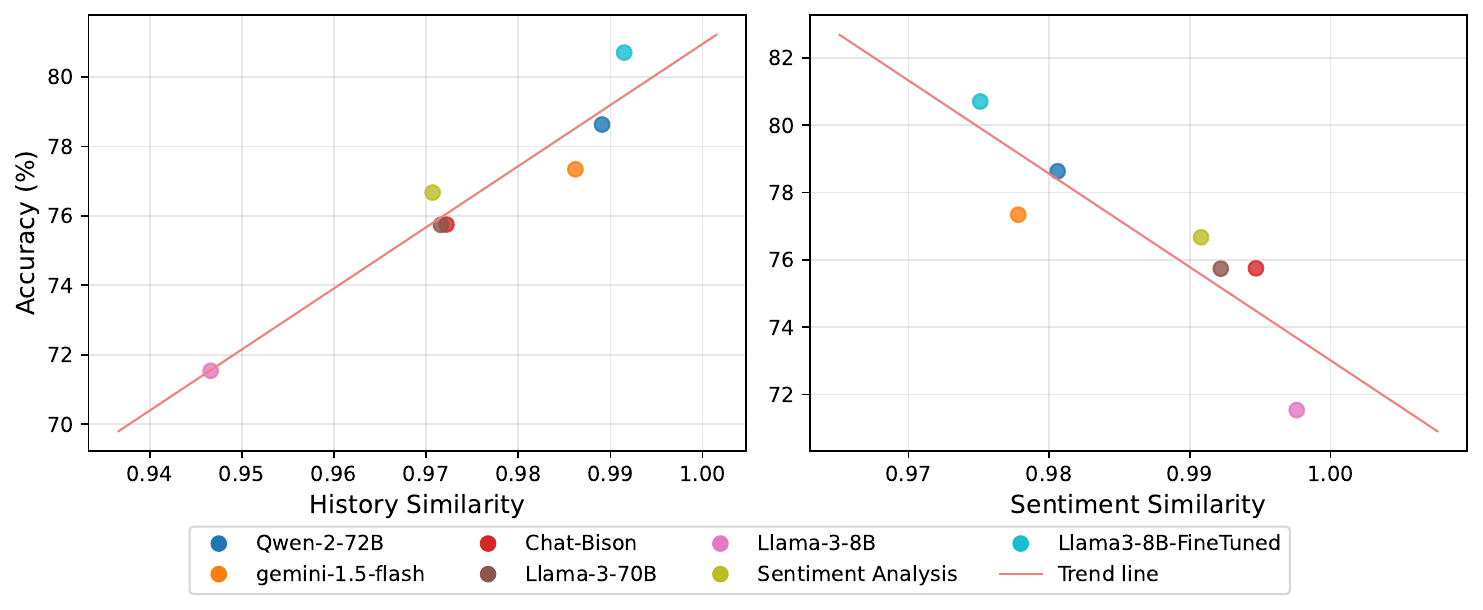}
  \caption{Relationship between similarity to human data and predictive performance of a model trained with LLM players from the same LLM. Left: grouping by sentiment shows an inverse trend between similarity and accuracy. Right: grouping by history shows a positive trend.
  }
\label{fig:history_correlations}
\end{figure}

Figure~\ref{fig:history_correlations} (Left) shows a direct relationship between similarity to human data (grouped by history) and the performance of the prediction model trained with the same data. 
Figure~\ref{fig:history_correlations} (Right) shows an inverse relationship between similarity to human data (grouped by sentiment) and the performance of a prediction model trained with the same data.
Note that this apparent linearity is likely an artifact of observing only a narrow range of high-performing models. Since the models tested are already trained to be semantically similar to humans, we are viewing a small, specific window of behavior. \footnote{To illustrate the broader relationship, consider a hypothetical model that generates random data: it would have a semantic similarity of 0 and predictive performance at a random-chance level. This suggests the true relationship across the entire spectrum of possible model behaviors is not linear.}
Taken together, both panels of Figure~\ref{fig:history_correlations} indicate that whereas similarity to human actions when grouped by \emph{history} is positively correlated with the predictive power of models trained on this data, similarity to human actions when grouped by \emph{sentiment} actually correlates negatively with it. These opposing trends offer a crucial insight into the nature of our task. The positive correlation between history-based similarity and predictive accuracy indicates that models capturing the strategic, sequential aspects of the game are more successful at forecasting human actions. Conversely, the negative correlation for sentiment-based similarity suggests that while LLMs can emulate human-like sentiment, this capability does not translate to, and may even detract from, predicting strategic choices. Our results, therefore, support the hypothesis that this is fundamentally a task of strategic decision-making, where game history is the dominant factor, rather than one of sentiment analysis.

\subsection{LLM as a Markov Model}\label{subsec:markov-llm}
The preceding analysis in the previous subsection established a crucial insight: the predictive power of LLM-generated data is strongly correlated with its ability to replicate human decision patterns based on interaction history, not just on the sentiment of the current message. This finding highlights that a successful data generator must model the strategic, history-dependent nature of human choice.

To further probe this insight, this subsection transitions from correlational analysis to a direct investigation into the importance of historical context. We ask: how does the length of the interaction history available to an LLM affect the quality of the synthetic data it produces? To address this, we will now formalize the role of history by modeling the LLM player as a system with a variable-length memory, allowing us to measure the impact of historical context on predictive accuracy systematically.

Our experimental setup approximates a Markov process: during both fine-tuning and synthetic data generation, it observes only the last $T$ rounds of its \emph{own} game history. To isolate the effect of the provided history from the model's pretrained world knowledge, we use Llama-3-8B, our weakest data generator. The rationale is that a smaller model possesses less world knowledge, forcing it to rely more heavily on the explicit history provided. This approach, however, did not improve prediction accuracy and even underperformed a simple language model baseline.

\begin{figure}[h]
  \centering \includegraphics[width=0.65\linewidth]{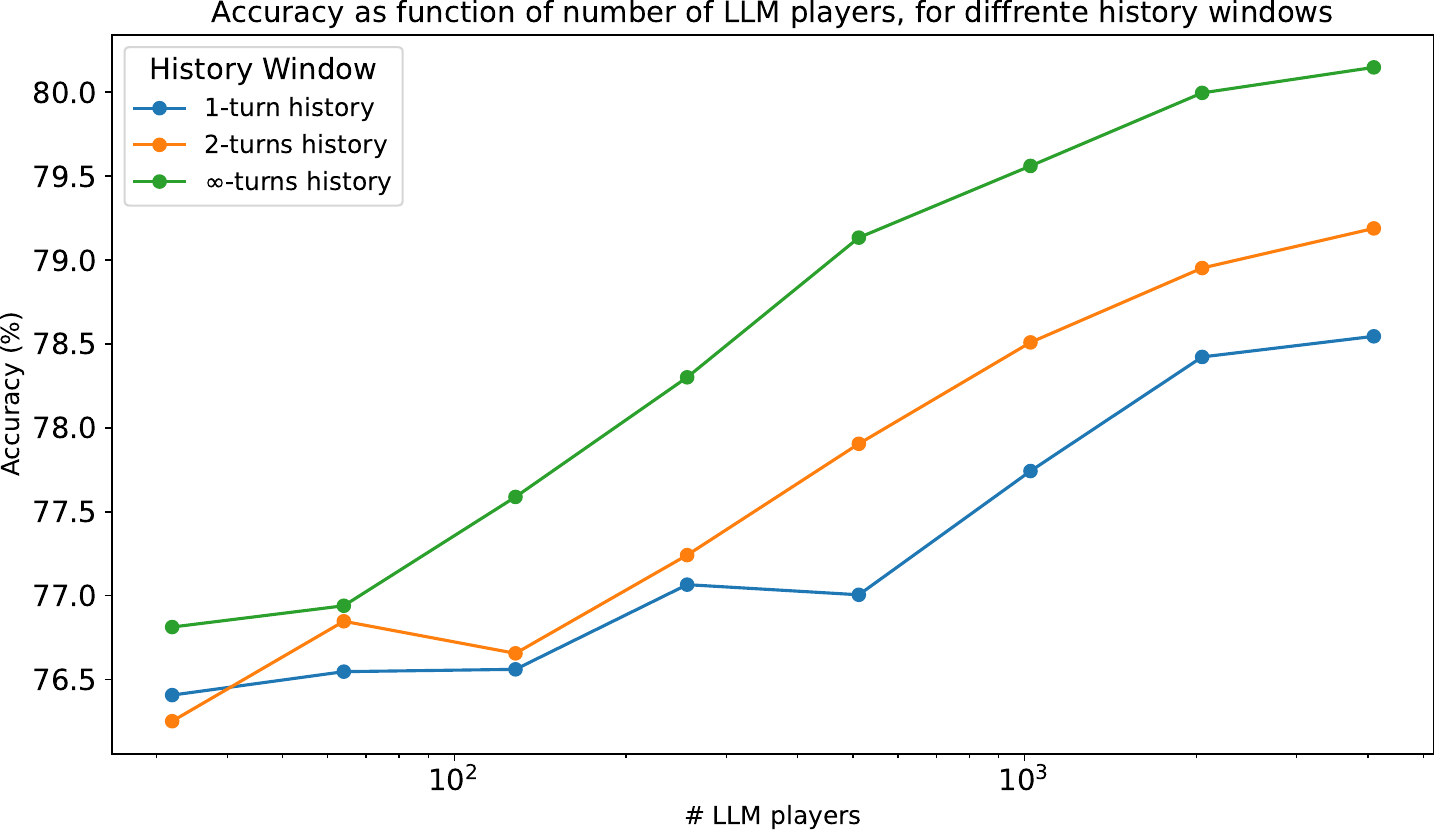}
  \caption{Predictive performance when training on data generated by \textsc{FT-Llama} tuned with varibale history lengths. Shorter visible history degrades downstream utility.
  }
\label{fig:markov_llm}
\end{figure}

Figure~\ref{fig:markov_llm} reports the performance of a predictive model trained on data generated by \textsc{FT-Llama-$T$-History} for $T \in \{1, 2, \infty\}$. The pattern is clear: the shorter the history visible to the LLM, the less useful the resulting synthetic data is for training an LSTM that predicts human behavior. These findings highlight that human-like predictive behavior cannot emerge from shallow, short-term dependencies alone.  As the LLM's visible history is truncated, the resulting synthetic data becomes less useful for training predictive models, reflected in their decreased accuracy. This performance drop suggests that the generated data is less faithful to the longer-term strategic dynamics present in human behavior, which are crucial for effective prediction and require access to a sufficiently long temporal context - capturing not only immediate reactions but also patterns that evolve across multiple interactions. In other words, the “memory” of past rounds plays a fundamental role in aligning synthetic agents with human decision processes.

\section{Predicting Behavior in Other Persuasion Games}
\label{sec:general_persuasion}

So far, we have focused on a specific setup of persuasion games. 
In this section, we ask a broader question: 
\textit{Can we predict human behavior in other language-based persuasion games?} 
While our previous analysis centered around a particular framing and mechanism of persuasion 
(e.g., the hotel setup and rule-based sender strategies), 
we now extend our investigation to \textbf{GLEE}, a different persuasion framework introduced by ~\citet{shapira2024glee}. 
\textbf{GLEE} is a two-player economic game framework where players communicate in natural language. In contrast to the single, narrative-driven scenario of the hotels game, GLEE defines a wide variety of abstract persuasion games by programmatically define by defining their structure and dynamics. It uses seven key parameters to do this, controlling factors such as the players' potential payoffs, the degree of conflict in their interests, and the information asymmetry between them.
By systematically varying these parameters, \citet{shapira2024glee} used the GLEE framework in order to generate a dataset of 360 distinct persuasion-game configurations, 
each representing a unique interaction between large language models. 
Table~\ref{tab:persuasion_configs} summarizes the seven parameters and the corresponding values 
appearing in this dataset.

\begin{table}[h]
    \centering
    \caption{Parameters defining the persuasion games that constitute the GLEE dataset.
The first four parameters ($p$, $v$, $M$, and $T$) can, in principle, take on infinitely many values, but here we report the discrete settings used to generate the GLEE dataset. The remaining parameters specify structural aspects of the game. In our experiment, we focus on the configurations with \textit{Messages type} is \textit{Textual} and \textit{Buyer type} is \textit{Long‐living}, leading to a total of 90 distinct experimental configurations.}
    \begin{tabular}{lll}
        \toprule
        \textbf{Parameter} & \textbf{Description} & \textbf{Values} \\
        \midrule
        $p$ & Prior probability that the product is good & $\tfrac13,\;0.5,\;0.8$ \\
        $v$ & Value of the high‐quality product for the buyer & $1.2,\;1.25,\;2,\;3,\;4$ \\
        $M$ & Scale parameter & $10^2,\;10^4,\;10^6$ \\
        $T$ & Number of rounds in the game & $20$ \\
        Complete Information & Does the seller know the buyer's value? & True, False \\
        Messages type & Whether free‐form textual messages are allowed & Binary, Textual \\
        Buyer type & Agent horizon & Long‐living, Myopic \\
        \bottomrule
    \end{tabular}
    \label{tab:persuasion_configs}
\end{table}

Among the 360 configurations in the GLEE dataset, we focus on the subset of persuasion games that allow us to study the strategic use of natural language and the development of reputation in repeated interactions. Specifically, we select configurations that satisfy the following two conditions:
\begin{itemize}
    \item The signal exchanged between players is a linguistic message - a natural language recommendation from the sender to the receiver about the quality of a product (\textit{Message type} is \textit{textual}).
    \item The interaction between players is repeated: the same sender and receiver play 20 consecutive rounds (\textit{Buyer type} is \textit{Long‐living}).
\end{itemize}

These conditions leave us with 90 persuasion configurations. Of these, 18 include data from games with human receivers, while the remaining 72 configurations feature data from games where both the sender and receiver were LLMs.
For each of these 90 configurations, we collected data from at least 35 games played between pairs of LLMs. For the 18 configurations that also involved human participants, we collected data from at least five additional games played with human receivers.

In this experiment, the sender's messages are freely generated in natural language, unlike in the hotels game, where the sender's messages were selected from a predefined set of Booking.com reviews. The EF inventory we use in our main prediction task is deliberately tailored to that setting: the features assume that each message can be decomposed into positive and negative parts, refer to a predefined list of hotel-related topics, and are defined with respect to a small, fixed set of scored reviews associated with each hotel. In GLEE's persuasion games, none of these structural assumptions hold: messages are unconstrained, free-form utterances written by humans or LLMs; they need not contain explicit positive/negative sections; and they are not tied to a particular scored review.
Instead, we use \textbf{TabSTAR}~\citep{arazi2025tabstar}, 
a tabular foundation model that uses an unfrozen text encoder and target-aware tokens to unify text and categorical data, enabling transfer across tasks and achieving state-of-the-art results on classification tasks involving tabular data with textual fields. 
The textual inputs include the sender’s message in the current and previous rounds, 
while the tabular features include the players’ previous decisions and payoffs, 
as well as the game’s parameterization. 
We also use \textbf{XGBoost}, 
which was previously employed in Section~\ref{sec:main res} 
and is known to be a strong baseline for such tasks~\citep{shwartz2022tabular}.

To predict the behavior of human receivers in these GLEE persuasion games, we trained the models on the full set of 3,150 games played between pairs of LLMs (35 games for each of the 90 configurations). We then tested these models on the decisions made across the 90 games involving human receivers.

As a human-data baseline, we trained separate models using a 5-fold cross-validation setup on the human receiver data. In each fold, we trained a model on 72 games (four out of the five games available for each of the 18 configurations with human data) and tested it on the remaining 18 games. This process ensures that predictions are generated for all 90 human-receiver games.
The GLEE dataset is highly imbalanced, featuring 11.8 positive examples for every negative one. This imbalance, which is significantly more pronounced than the 1.4-to-1 ratio in the hotels game dataset, makes standard accuracy an unsuitable performance measure. Consequently, we use the Area Under the ROC Curve (AUC-ROC) as a more appropriate evaluation metric.

\begin{figure}[h]
  \centering \includegraphics[width=\linewidth]{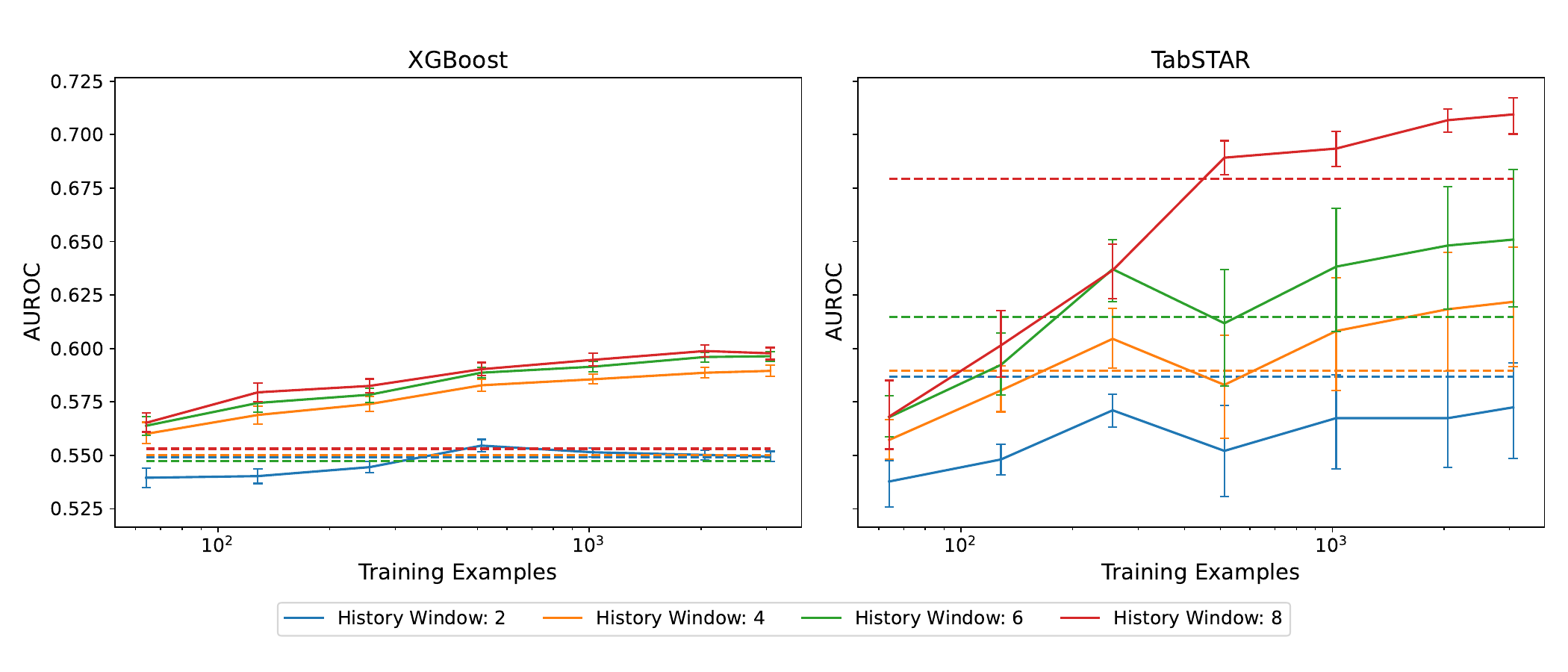}
  \caption{AUC-ROC of the XGBoost model (left) and the TabSTAR model (right), trained on datasets containing different numbers of games between pairs of LLMs (x-axis). The color indicates the length of the interaction history used as input to the prediction model. The dashed lines represent the performance of models trained on data from 72 games between an LLM sender and a human receiver.
  }
\label{fig:glee_auc}
\end{figure}

Figure~\ref{fig:glee_auc} presents the performance of XGBoost (left) and TabSTAR (right)
in predicting human receiver decisions. 
Each model was trained on datasets containing between 64 and 3072 games (x-axis), 
using histories of length $T \in \{2, 4, 6, 8\}$ rounds. 
For $T \in \{4, 6, 8\}$, models trained on data from the 1024 LLM-vs-LLM games 
significantly outperformed models trained on data from the 72 LLM-vs-Human games, 
with TabSTAR outperforming XGBoost across all configurations. 
These results demonstrate that predicting human decisions in persuasion games using LLM-generated data 
is not limited to specific setups such as the hotels game, but generalizes to broader persuasion frameworks like the ones in GLEE.

We note that the comparison between 1024  LLM-vs-LLM games and 72 LLM-human games is chosen to illustrate a practical trade-off. The 72 games represent the extent of the available, yet costly to obtain, human training data (4 games for each of the 18 configurations). In contrast, generating synthetic data from LLM-vs-LLM interactions is computationally inexpensive and highly scalable. The 1024-game threshold is not an arbitrary hyperparameter but an empirical finding from our scaling experiments; it is the point at which the volume of inexpensive synthetic data compensates for its lower fidelity, outperforming the model trained on the limited set of real human interactions. This demonstrates a general principle: when human data is scarce, it can be more effective to train models on a much larger corpus of synthetically generated data.

\section{Discussion}
\label{sec:disc}

\paragraph{Conclusion}
The main goal of this paper is to illustrate a use case of using LLM-generated data for training a prediction model for human choices in a language-based persuasion task. We built upon the language-based persuasion game and showed that training a choice prediction model on a dataset containing no human choice data at all can even outperform the same model trained on an actual human-generated dataset, given enough generated data points. This observation has major implications for understanding synthetic data potential in the context of enhancing human choice prediction. 

We have compared the results obtained by our approach to a naive sentiment-analysis baseline, in which decisions of synthetic players are made solely based on the current hotel review, without considering the entire history of interaction between the DM and the expert. The fact that our method significantly outperforms the baseline highlights the importance of generating synthetic decision data in a way that captures strategic behavior.

We then demonstrated the effectiveness of combining LLM-generated data with existing human data to further enhance predictive power. To better understand and characterize the potential and limitations of our LLM-based approach, we also provided a per-expert analysis and evaluated the LLM-based approach against each expert strategy separately.
On the negative side, we showed that our LLM-based approach leads to lower calibration compared to predictors trained on human data, giving rise to an accuracy-calibration tradeoff.

We further explored alternative roles for LLMs in the prediction pipeline, showing that they can serve not only as data generators but also directly as predictors. Fine-tuning improved performance in both stages, and utilizing human data for both generation and classification --- a strategy we termed \emph{Double Use of human data for Augmented Learning (DUAL)} --- offered additional gains in terms of calibration.

Using our framework, we analyzed behavioral patterns in a language-based persuasion game and showed that decision-making was shaped by both the linguistic signal and the history of past interactions. Capturing this historical dimension was crucial for predictive accuracy: models that accounted for history outperformed those relying on short interaction horizons or purely linguistic signals. Finally, we provided preliminary evidence that the effectiveness of LLM-generated data extends beyond a single experimental setting, supporting the broader applicability and reliability of our approach.

While the findings of this paper are specific to our experimental context, they offer a novel approach to studying and predicting human behavior. Future research may focus on exploring the predictive power of LLM-generated data beyond the context of language-based persuasion games, as well as characterizing the boundaries and limitations of this approach, e.g., by utilizing explainability methods to better understand the difference between models trained on different data sources.

\paragraph{Limitations and future work}

This work suffers from several limitations, giving rise to potential future work. First, it focuses on a specific class of games, namely language-based persuasion games. While these games have major importance both in the economics literature and in the NLP community, the fact that the approach of utilizing LLMs for training data generation is demonstrated solely in this setup is indeed restrictive. We view this work as a first attempt to apply the approach in a complex and realistic economic setup, which is also grounded in a particular real world application. Focusing on this particular setup also enables a richer analysis of the results and discussion of the assumptions. For instance, discussing the specific persuasion strategies considered and the evaluation with respect to each strategy separately, is a kind of contribution that is particularly relevant in the context of persuasion games. We hope that this contribution will encourage the community to consider different applications of the proposed approach, as well as further suggest extensions and improvements.

Second, even in the context of persuasion games, the restriction to a specific set of expert strategies is limiting by nature. We highlight that this limitation comes from a practical argument of balancing budget constraints and the need for collecting a large enough sample for each strategy, as well as the need for avoiding the collection of another human choice dataset in addition to the dataset of \citet{shapira2025human}, as doing so may impose non trivial challenges in equalizing the conditions among human participants. These constraints indeed call for taking a critical view of the particular strategies considered, and we believe that enriching the discussion on the extent to which these strategies are relevant and representative can be viewed as a major contribution. Here we suggest two alternative justifications for the particular set of strategies considered: (1) an intuitive behavioral interpretation for each strategy; and (2) a classification of all strategies to types covering the entire strategy space (\emph{naive}, \emph{stationary} and \emph{adaptive}), where each type is indeed represented in our strategies set (see Appendix \ref{app:experts}). We acknowledge the need for continuously discussing, questioning and criticizing the limitations in the selection of the strategies that define the human choice prediction task.

Another limitation, which may be more application-specific, is the mis-calibration of prediction models trained solely on LLM-generated data, which is studied and discussed at the Section \ref{sec:calibration}. We highlight that raising awareness of this downside of the proposed approach is an interesting insight on its own, and it raises some potential future questions regarding the connections between LLMs and calibration that may be of interest to the NLP community. However, we show that DUAL, or adding human data to a model trained using LLM-generated data, can potentially mitigate the mis-calibration effect.

Lastly, a key question that remains unanswered is \emph{when} and \emph{why} the LLM-based approach is less effective. Namely, can we characterize those specific cases (e.g., specific expert strategies, or even specific decisions) in which models trained on LLM-generated data fall short? One can view our per-expert analysis (Section \ref{sec:per expert main res}) as a first step towards this end, yet the complete characterization is left as an interesting future question.

\paragraph{Ethical considerations}

This work may also have several ethical implications.
In behavioral economics studies, ethical issues concern keeping participants' privacy, and this work suggests ways to make that process easier. From this perspective, our suggested approach offers a solution for such ethical considerations involved with experimental economics.
On the other hand, this very same approach, which serves effective human choice prediction, has the potential to be utilized maliciously. The potential of using LLMs to enhance human choice prediction, demonstrated in this work, calls for clear ethical guidelines to safeguard against its potential for harm, emphasizing the importance of responsible use.

\section*{Acknowledgements}
R. Reichart and E. Shapira have been partially supported by a VATAT grant on data science. The work of O. Madmon, E. Shapira and M. Tennenholtz is funded by the European Research Council (ERC) under the European Union’s Horizon 2020 research and innovation program (grant agreement n° 740435). We would like to thank Itamar Reinman for providing valuable comments on an earlier version of this paper.

%% file: tables/llm_dataset.tex
\begin{tabular}{ll}
\toprule
\textbf{Persona Type} & \textbf{Prompt Prefix} \\
\midrule
Optimistic & "Behave like an optimistic person." \\ 
Pessimistic & "Behave like a pessimistic person." \\ 
Price & "Behave like a person to whom the hotel's price is important." \\ 
Facilities & "Behave like a person who values the facilities offered by the hotel." \\ 
Room & "Behave like a person who cares about the quality of the room in the hotel." \\ 
Location & "Behave like a person for whom the location of the hotel is important." \\ 
Staff & "Behave like a person who cares about the treatment they will receive \\ & from the hotel staff." \\ 
Sanitary & "Behave like a person to whom the sanitary conditions of the hotel are \\ & important." \\ 
\bottomrule
\end{tabular}

%% file: tables/datasets_statistics.tex
\begin{table}[h]
    \caption{
Number of players created by each LLMs.}
\begin{tabular}{lccccc}
        \toprule
        \textbf{Persona} & \textbf{Qwen-2-72B} & \textbf{Llama-3-8B} & \textbf{Llama-3-70B} & \textbf{gemini-1.5-flash} & \textbf{Chat-Bison} \\
        \midrule
        Optimistic & 519 & 512 & 208 & 129 & 514 \\
        Pessimistic & 528 & 512 & 208 & 144 & 512 \\
        Price & 538 & 512 & 192 & 128 & 514 \\
        Facilities & 525 & 517 & 208 & 128 & 514 \\
        Room & 513 & 512 & 145 & 138 & 514 \\
        Location & 512 & 516 & 193 & 132 & 514 \\
        Staff & 512 & 528 & 208 & 128 & 514 \\
        Sanitary & 530 & 512 & 208 & 129 & 514 \\
        \midrule
        All & 4177 & 4121 & 1570 & 1056 & 4110 \\
        \bottomrule
    \end{tabular}
    \label{tab:dataset_statistics}
\end{table}

%% file: figures/personas_effect_code.tex
\begin{figure}[h]
  \centering \includegraphics[width=300pt]{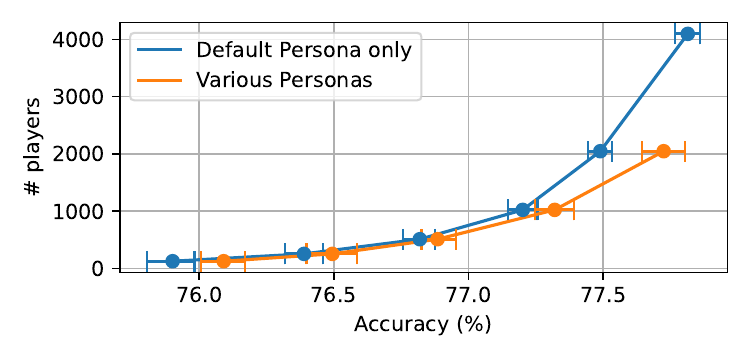}
  \caption{Number of LLM-generated players (using Chat-Bison) required for achieving different levels of accuracy (with the LSTM predictor), with persona diversification (various personas) and without persona diversification ('default' persona only). Interestingly, note that the higher the desired accuracy, the larger the gap between the required sample sizes of the two methods.}
\label{fig:personas_effect}
\end{figure}

%% file: figures/main_result_code.tex
\begin{figure*}[t]
  \centering
  \includegraphics[width=\textwidth]{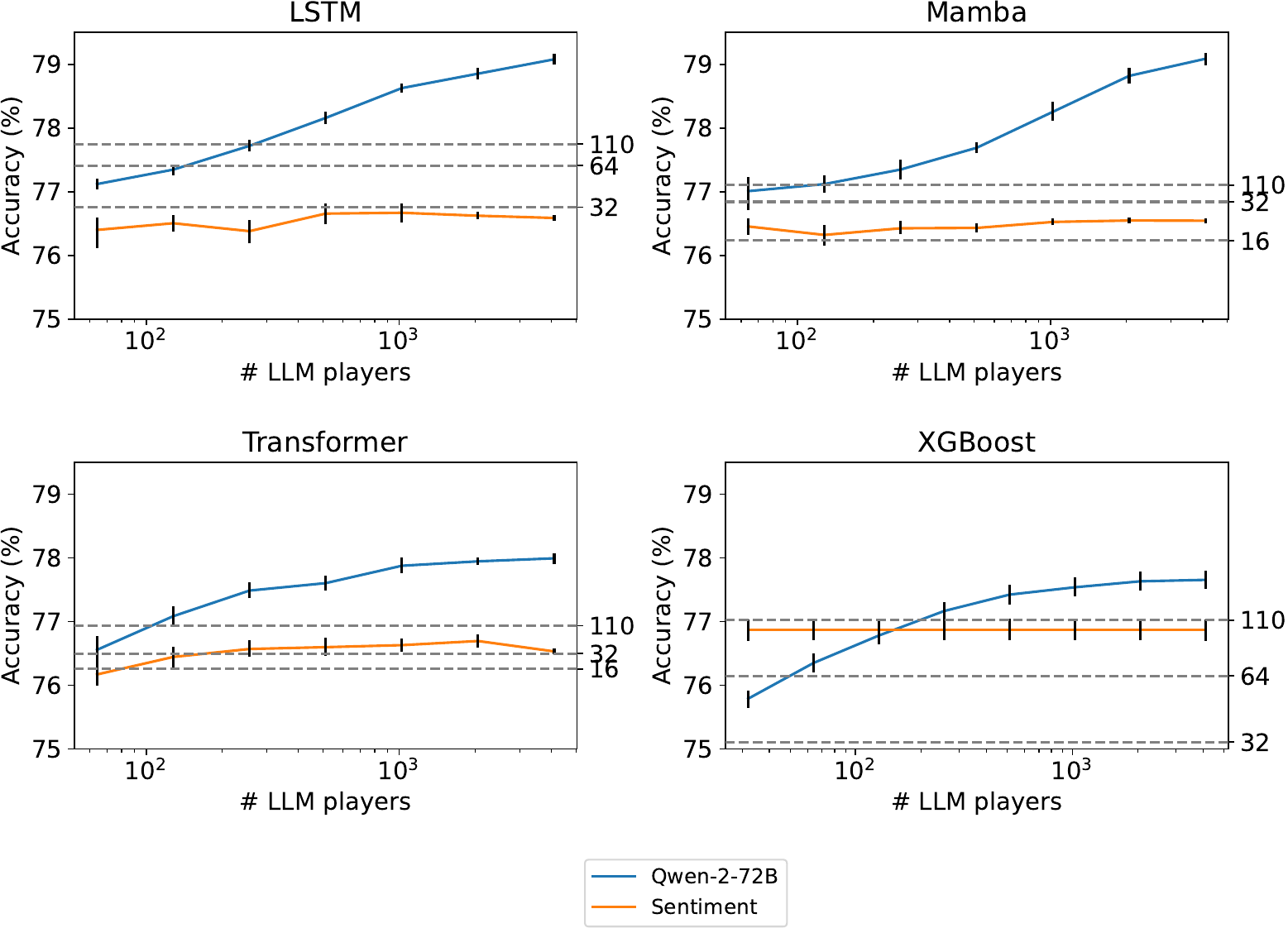}
  \caption{Accuracy obtained by prediction models trained on different data sources.
  Grey lines represent the accuracy obtained by a model trained on human data with a different number of players. 
  Results are shown for LSTM, transformer, Mamba and XGBoost.
  Notably, for all prediction models training on LLM-generated data outperforms training on actual human choice data when the number of LLM players is large enough. In addition, the LLM-based training paradigm outperforms the sentiment analysis baseline, implying that allowing simulated players to determine behavior (and not just to interpret the textual signal) yields a better predictor.}
\label{fig:main_result}
\end{figure*}

%% file: tables/main_results.tex
\begin{table}[h]
    \centering
    \caption{
Accuracy obtained by LLM datasets for varying training sizes, along with the Sentiment baseline, for different numbers of players in the training set.
Configurations that provide better accuracy than a model trained with data of 16 humans are \underline{underlined}.
Configurations that provide better accuracy than a model trained with data of 110 humans are \textbf{\underline{both underlined and bolded}}. Some of the values here are missing due to the high cost of generating LLM players.
}
\begin{tabular}{lccccccc}
        \toprule
        \textbf{Training Size} & \textbf{64} & \textbf{128} & \textbf{256} & \textbf{512} & \textbf{1024} & \textbf{2048} & \textbf{4096} \\
        \midrule
        Qwen-2-72B & \underline{77.12} & \underline{77.35} & \underline{77.72} & \textbf{\underline{78.16}} & \textbf{\underline{78.63}} & \textbf{\underline{78.85}} & \textbf{\underline{79.08}} \\
        gemini-1.5-flash & \underline{76.13} & \underline{76.30} & \underline{76.67} & \underline{77.16} & \underline{77.34} & - & - \\
        Chat-Bison & 73.65 & \underline{74.12} & \underline{74.72} & \underline{75.34} & \underline{75.75} & \underline{75.93} & \underline{76.04} \\
        Llama-3-70B & \underline{75.84} & \underline{75.71} & \underline{75.53} & \underline{75.42} & \underline{75.74} & - & - \\
        Llama-3-8B & 72.04 & 71.45 & 71.72 & 71.38 & 71.54 & 71.70 & 71.93 \\
        \midrule
        Sentiment & \underline{76.40} & \underline{76.51} & \underline{76.38} & \underline{76.66} & \underline{76.67} & \underline{76.62} & \underline{76.59} \\
        \bottomrule
    \end{tabular}
    \label{tab:offline_simulation}
\end{table}

%% file: figures/human_and_llm_code.tex
\begin{figure*}[t]
  \centering \includegraphics[width=1\textwidth]{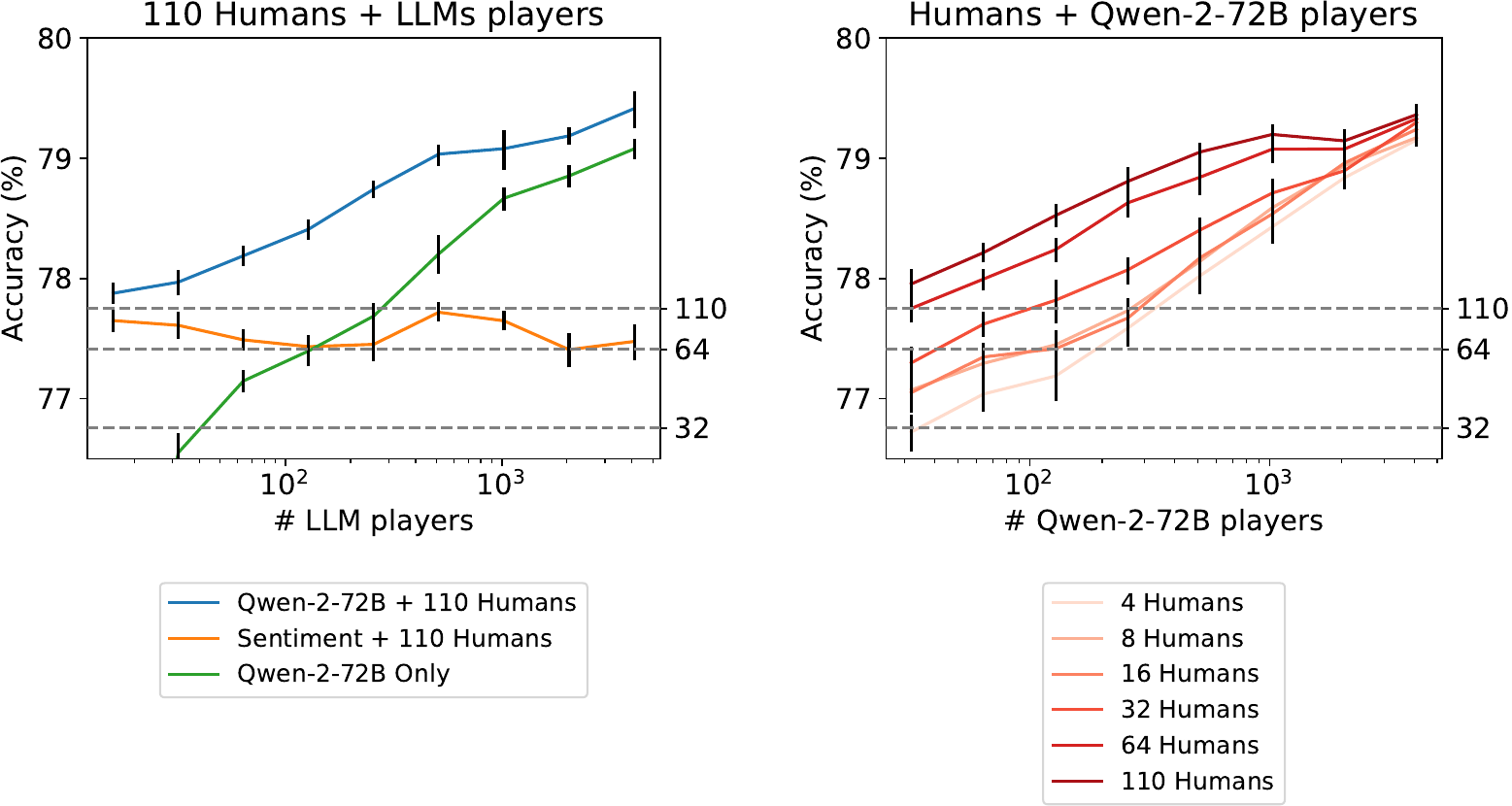}
  \caption{\textbf{Left:} Accuracy of models trained with data containing 110 humans and a varying number of LLM players (x-axis). \textbf{Right:} Accuracy of a model trained with data containing a varying number of humans (color) and a varying number of players generated by Qwen-2-72B (x-axis). Gray lines: Models trained with data comprised of only human players. Combining human and LLM players outperforms training solely on human players.}
\label{fig:human_and_llm}
\end{figure*}

%% file: figures/two_strategies_performance_code.tex

\begin{figure*}[t]
  \centering \includegraphics[width=1\textwidth]{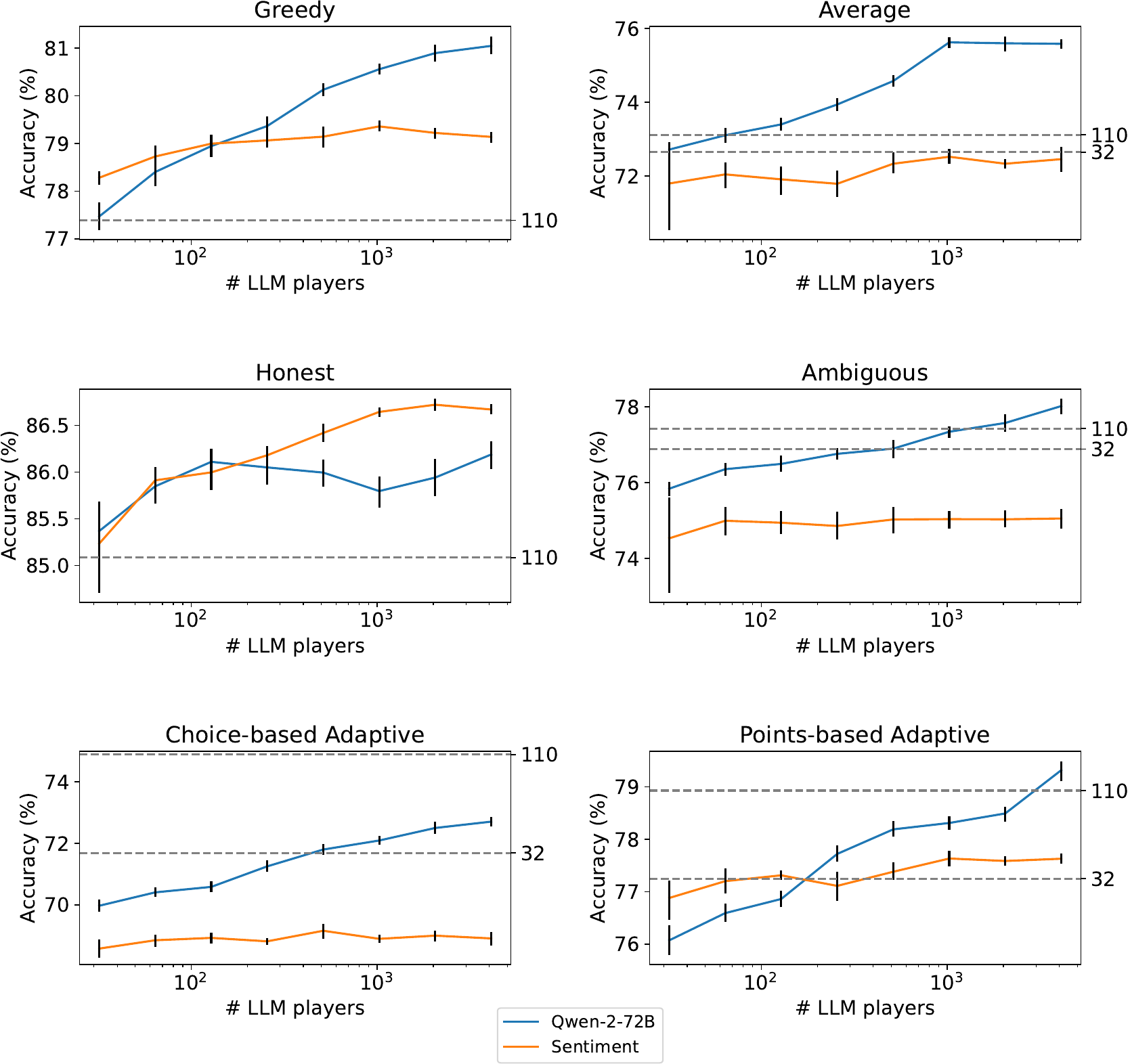}
  \caption{Accuracy on individual expert interactions, obtained by prediction models for the three training configurations: human data, LLM-generated data, and the sentiment baseline. Prediction models trained on LLM-generated data outperform those trained on 32 human players and almost always surpass those trained on 110 human players.}
\label{fig:strategies_performance}
\end{figure*}

%% file: appendix.tex
\appendix

\section{Expert Strategies}
\label{app:experts}

Our paper is concerned with solving the task of predicting human decisions in a language-based persuasion game, as studied by \citet{apel2022predicting}. \citet{shapira2025human} collected and published a dataset of human-bot interactions, in which human players interact with six expert bots. We use these interactions to define the prediction task, hence we use the same expert strategies to collect our LLM dataset. We devote this section to discussing the six strategies introduced by \citet{shapira2025human} and claiming that this set is somewhat representative of the entire strategy space of the expert in the persuasion game. Figure \ref{fig:all_experts} provides the binary tree representation of all six expert strategies.

\begin{table*}[h]
\renewcommand{\arraystretch}{1.5}
\centering
\begin{tabular}{cc}
\begin{minipage}{.5\textwidth}
\centering
\begin{tikzpicture}[
        level 1/.style={sibling distance=35mm},
        level 2/.style={sibling distance=35mm},
        every node/.style={draw, rectangle, align=center},
        edge from parent/.style={draw, -latex},
        edge from parent path={(\tikzparentnode.south) -- (\tikzchildnode.north)}
    ]
    \node {Send review with \\the highest score};
\end{tikzpicture} \\
\end{minipage} &
\begin{minipage}{.5\textwidth}
\centering
\begin{tikzpicture}[
        level 1/.style={sibling distance=35mm},
        level 2/.style={sibling distance=35mm},
        every node/.style={draw, rectangle, align=center},
        edge from parent/.style={draw, -latex},
        edge from parent path={(\tikzparentnode.south) -- (\tikzchildnode.north)}
    ]
    \node {Send review whose\\ score is closest to\\the mean score};
\end{tikzpicture} \\
\end{minipage} \\
(A) \texttt{Greedy} strategy & (B) \texttt{Average} strategy \\
& \\
\begin{minipage}{.5\textwidth}
\centering
\begin{tikzpicture}[
        level 1/.style={sibling distance=35mm},
        level 2/.style={sibling distance=35mm},
        every node/.style={draw, rectangle, align=center},
        edge from parent/.style={draw, -latex},
        edge from parent path={(\tikzparentnode.south) -- (\tikzchildnode.north)}
    ]
    \node {Is the current hotel of high quality?}
        child {node {Choose review with \\ the highest score}
            edge from parent node[left, draw=none] {Yes}
        }
        child {node {Choose review with \\ the lowest score}
            edge from parent node[right, draw=none] {No}
        };
\end{tikzpicture}
\\ 
\end{minipage} &
\begin{minipage}{.5\textwidth}
\centering
\begin{tikzpicture}[
        level 1/.style={sibling distance=40mm},
        level 2/.style={sibling distance=35mm},
        every node/.style={draw, rectangle, align=center},
        edge from parent/.style={draw, -latex},
        edge from parent path={(\tikzparentnode.south) -- (\tikzchildnode.north)}
    ]
    \node {Is the current hotel of high quality?}
        child {node {Choose review with \\ the highest score}
            edge from parent node[left, draw=none] {Yes}
        }
        child {node {Choose review whose \\ score is closest to \\ the mean score}
            edge from parent node[right, draw=none] {No}
        };
\end{tikzpicture}
\\
\end{minipage} \\
(C) \texttt{Honest} strategy & (D) \texttt{Ambiguous} strategy  \\
& \\

\begin{minipage}{.5\textwidth}
\centering
\begin{tikzpicture}[
        level 1/.style={sibling distance=40mm, level distance=20mm},
        level 2/.style={sibling distance=35mm},
        every node/.style={draw, rectangle, align=center},
        edge from parent/.style={draw, -latex},
        edge from parent path={(\tikzparentnode.south) -- (\tikzchildnode.north)}
    ]
    \node {Did the DM go to the \\hotel in the previous round?}
        child {node {Choose review whose \\ score is closest to \\ the mean score}
            edge from parent node[right, draw=none] {Yes}
        }
        child {node {Choose review with \\ the highest score}
            edge from parent node[left, draw=none] {No}
        };
\end{tikzpicture}
\\ 
\end{minipage} &
\begin{minipage}{.5\textwidth}
\centering
    \begin{tikzpicture}[
        level 1/.style={sibling distance=45mm, level distance=20mm},
        level 2/.style={sibling distance=35mm, level distance=20mm},
        every node/.style={draw, rectangle, align=center, scale=0.75},
        edge from parent/.style={draw, -latex},
        edge from parent path={(\tikzparentnode.south) -- (\tikzchildnode.north)},
        scale=0.75
    ]
    \node {Is the current hotel of high quality?}
        child {node {Choose review whose \\ score is closest to \\ the mean score}
            edge from parent node[left, draw=none] {Yes}
        }
        child {node {Has the DM earned \\more points than you in \\all previous rounds?}
            child {node {Choose review with \\ the highest score}
                edge from parent node[left, draw=none] {Yes}
            }
            child {node {Choose review with \\ the lowest score}
                edge from parent node[right, draw=none] {No}
            }
            edge from parent node[right, draw=none] {No}
        };
    \end{tikzpicture}
\\ 
\end{minipage} \\
(E) \texttt{Choice-based Adaptive} strategy & (F) \texttt{Points-based Adaptive} strategy \\
\end{tabular}
\captionof{figure}{Expert strategies, represented as binary trees.}
\label{fig:all_experts}
\end{table*}

\paragraph{Strategies interpretation} Importantly, we argue that each of the six strategies has a clear and intuitive behavioral interpretation. The \texttt{Greedy}, briefly discussed in the introduction, is the strategy according to which the expert always reveals the most positive review. The \texttt{Average} strategy is a more careful strategy that always displays the (closest to the) average review, aiming to build trust by revealing more representative information to the DM.

Another strategy that aims to build trust with the DM is the \texttt{Honest} strategy, also discussed briefly in the introduction. In this strategy, the expert reveals the most positive review when the hotel is of high quality, but "warns" the DM when the hotel is of low quality by revealing the worst possible review. A similar, yet more sophisticated and manipulative strategy is the \texttt{Ambiguous} strategy. In this strategy, the expert also reveals the most positive review for good hotels, but when the hotel is bad she ambiguously selects a not-too-negative review. This principle of revealing the state when it is good and "bluffing" when it is bad is fundamental in the economic literature, and turns out to be an optimal sender strategy in economic setups that can be modeled as persuasion games, such as product adoption games \cite{arieli2023reputation}.

The last two strategies, \texttt{Choice-based Adaptive} and \texttt{Points-based Adaptive}, are slightly more sophisticated as they aim to adaptively control the behavior of the DM. The \texttt{Choice-based Adaptive} completely ignores the true quality of the hotel and only conditions the review selection rule on whether the DM opted-in in the previous round or not. This can be seen as trying to push the hotel more aggressively after failing to do so in the previous round. \texttt{Points-based Adaptive} follows a similar high-level principle, with two major differences: it first takes into account the actual quality of the hotel, and it selects the review to present based on the number of points gathered by the DM, and not by the selected action.

\paragraph{Strategies classification} We argue that these can be classified into three different strategy classes:

\begin{itemize}
    \item \textbf{Naive strategies.} These are strategies in which the expert's action is independent of both the actual quality of the hotel and the interaction history.
    \item \textbf{Stationary strategies.} These are strategies in which the expert's action depends solely on the actual quality of the hotel (but it is independent of the interaction history).
    \item \textbf{Adaptive strategies.} These are strategies in which the expert's action depends on the interaction history.
\end{itemize}

It is clear that \texttt{Greedy} and \texttt{Average} are naive, \texttt{Honest} and \texttt{Ambiguous} are stationary, and \texttt{Choice-based Adaptive} and \texttt{Points-based Adaptive} are adaptive. Trivially, every possible strategy is either naive, stationary, or adaptive. While the entire strategy space is infinite, we argue that the consideration of two strategies of each class within the experimental setup is somewhat representative.




\section{Prompts and Example Conversation}
\label{app:prompt-conv-example}

\newcommand\LLMmessage[1]{\textcolor{orange}{#1}}
\newcommand\Agentmessage[1]{\textcolor{cyan}{#1}}
\newcommand\Naturemessage[1]{\textcolor{violet}{#1}}

This appendix explains how we collected the data through interaction with an LLM that simulated a DM. In addition, we introduce the beginning of an interaction and its first two rounds.

The colors represent the message sender: messages from nature appear in \Naturemessage{purple}, messages from the Sender appear in \Agentmessage{blue}, and messages from the LLM appear in \LLMmessage{orange}.
Every round, the LLM-Chat agent receives a message containing all the purple and blue parts that have been sent since its last message.

\label{app:conversasion}
\begin{figure}[h]
  \centering
\includegraphics[width=220pt]{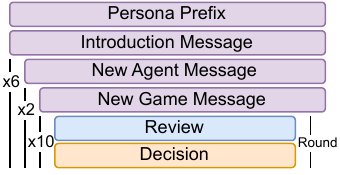}
  \caption{The data collection pipeline of a single LLM-based player. The player is initialized with a prompt that contains the persona prefix followed by the introduction message. Then, nature (purple) manages the conversation between the agent (blue) and the LLM-based player (orange). Practically, the LLM-Chat agent receives a concatenation of all purple and blue parts sent since its last message as a single message.}
\label{fig:conversation}
\end{figure}

\begin{quote}
\Naturemessage{
Behave like a person to whom the hotel's price is important. \\
Let's play a game!\\\#\#\#\\
Introduction: \\
Are you the vacation planner at your house? Think you always know how to choose the best hotel? Start to plan your 10-days trip with our travel agents. Just remember - they don't always want the best for you, and might have their own strategy to make you book the hotel they try to promote! Travel or Trouble is a strategy game in which you will try to outsmart our traveling agents and plan the perfect vacation for you. Each game consists of 10 rounds, in each round, one of our traveling agents will introduce you to a review for a new hotel they think might suit you, and you will have to choose: either book the hotel or stay home.\\
\#\#\#\\
The game: \\
You will play as the traveler and encounter several travel agents in the game. In each round, the agent will provide you with a message about the hotel, and you will decide whether to go to the hotel or stay at home. Your goal is to go to the good hotels and avoid the bad ones.\\
Meet your new travel agent: David! \\
You'll be playing the next game with David as your travel agent. \\
You need to earn 10 points to win the game.\\
\#\#\# \\
Round: 1 \textbar \space You have 0 points\\
David's review about the hotel: \\
\~\\Positive:}
\Agentmessage{ Location}\\
\Naturemessage{Negative:}
\Agentmessage{Bad hotel. Bad service! First problems with the doors keycard, had to go up 2 times before receptionist believed us. Then no light in the bathroom. Unable to fix..so to toilet in the dark. Compensation was free breakfast. Next morning, no water on top floor, clearly a pressure problem, new receptionist had no cloe, "new problem, never had complaints". Then he wanted us to pay for breakfast :(  I'm not a complainer but this was bad! Do not get fooled by rating..hotel without a soul, and nobody who takes responability for it.\\}
\Naturemessage{\~\\
Choose your action: [ Go \textbar \space Don't Go ]}\\
\LLMmessage{Go} \\
\Naturemessage{Round results: \\
This hotel is bad, You should have skipped it. \\
This round, you earn no points. \\
\#\#\#\\
Round: 2 \textbar \space You have 0 points\\
David's review about the hotel: \\
\~\\
Positive:}
\Agentmessage{ Big and spacious.  This apartment was EXCEPTIONALLY CLEAN and a great value in Brussels.  If given the chance to stay here again I would not hesitate to do so.  The staff was also very nice.}\\
\Naturemessage{Negative:}
\Agentmessage{ I have nothing negative to say about our experience.\\
}
\Naturemessage{\~\\Choose your action: [ Go \textbar \space Don't Go ]}\\
\LLMmessage{Go}
\end{quote}

\section{Review-Score Distribution Extraction for the Linguistic Baseline Method}
\label{app:review2score}

This appendix explains how we extract the score distribution induced by the LLM for a given review.
We use this distribution in the baseline method, as explained in section \ref{sec:lig_baseline}.
We asked the language model to transform a review into a numerical score, and extracted the underlying score distribution from the distribution the model assigns to the different numerical tokens.

For instance, if upon observing review $r$ the model assigns a probability of 0.4 to the first token of the output to 
be 8, then $P (8 \leq s < 9|r) = 0.4$.

\newcommand\dynamic[1]{\textcolor{red}{#1}}
The following text describes the format prompt we used for extracting the probabilities. The \dynamic{red} parts represent the review itself. 
\begin{quote}
Rank the value of the hotel as presented by the review, from 1 to 100, with 80 being the minimum score for a hotel you would like to stay in. \\
    Positive: \dynamic{Big and spacious.  This apartment was EXCEPTIONALLY CLEAN and a great value in Brussels.  If given the chance to stay here again I would not hesitate to do so.  The staff was also very nice.} \\ 
    Negative: \dynamic{I have nothing negative to say about our experience.} \\
    Answer only with your value!
\end{quote}

\section{Strategy Evaluation From the Expert's Perspective}
\label{app:per-exp}

\begin{figure*}[h]
  \centering
\includegraphics[width=\linewidth]{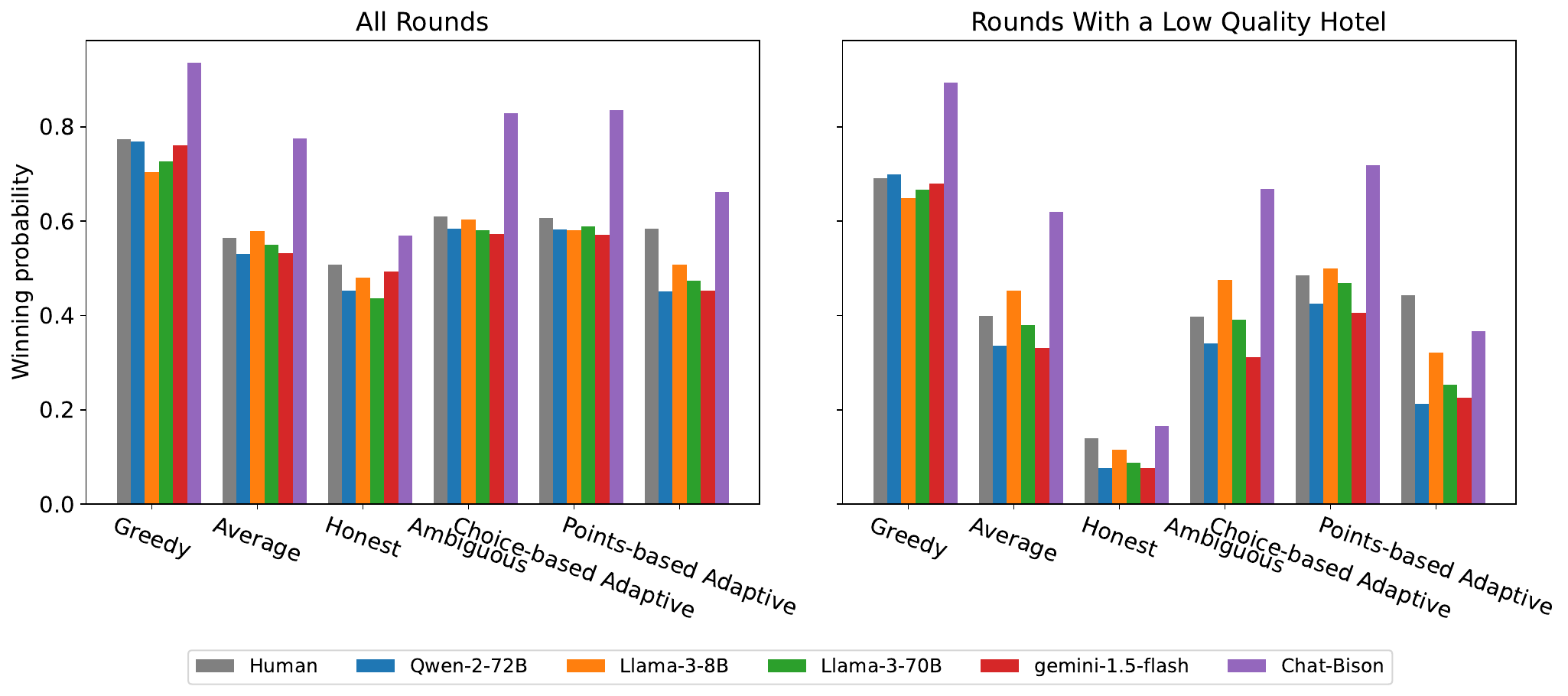}
  \caption{Winning rates of the expert against both LLM-based DMs and human DMs for each expert strategy.}
\label{fig:winning_percentage}
\end{figure*}

In this appendix, we evaluate the different strategies from the expert's perspective: Figure \ref{fig:winning_percentage} (Left) shows the expert's winning rate for each expert strategy, against human players and the different LLM players. Figure \ref{fig:winning_percentage} (Right) shows the same metric \emph{conditional on the actual hotel's quality being low.} That is, the latter shows the persuasion power of each expert in cases where the DM should not opt-in.

A first observation is that most LLM players behave similarly to human DMs. The only exception is Chat-Bison players, which seem to be significantly easier to manipulate.
Additionally, as one could expect, for all players, the opt-in frequency is significantly higher when we do not condition the hotel as being of low quality.
In terms of evaluating the strategies from the expert's perspective, it is notable that \texttt{Greedy} is indeed the best strategy for the expert, which is also consistent with \citet{raifer-etal-2022-designing}.

\section{All Possible Histories}
\label{app:history_options}
Table~\ref{tab:history-grouping-states} describes the 21 possible states, where the data is grouped by history for calculating similarity. 

\begin{table}[h]
\centering
\caption{Full list of the 21 history states used by the \emph{History grouping}.}
\label{tab:history-grouping-states}
\setlength{\tabcolsep}{8pt}
\begin{tabular}{clllll}
\toprule
\textbf{ID} & \textbf{Round} & \textbf{last\_didGo} & \textbf{last\_didWin} & \textbf{last\_last\_didGo} & \textbf{last\_last\_didWin} \\
\midrule
1  & 1   & N/A           & N/A            & N/A            & N/A \\
\midrule
2  & 2   & \texttt{True}  & \texttt{True}  & N/A            & N/A \\
3  & 2   & \texttt{True}  & \texttt{False} & N/A            & N/A \\
4  & 2   & \texttt{False} & \texttt{True}  & N/A            & N/A \\
5  & 2   & \texttt{False} & \texttt{False} & N/A            & N/A \\
\midrule
6  & 3+  & \texttt{True}  & \texttt{True}  & \texttt{True}  & \texttt{True}  \\
7  & 3+  & \texttt{True}  & \texttt{True}  & \texttt{True}  & \texttt{False} \\
8  & 3+  & \texttt{True}  & \texttt{True}  & \texttt{False} & \texttt{True}  \\
9  & 3+  & \texttt{True}  & \texttt{True}  & \texttt{False} & \texttt{False} \\
10 & 3+  & \texttt{True}  & \texttt{False} & \texttt{True}  & \texttt{True}  \\
11 & 3+  & \texttt{True}  & \texttt{False} & \texttt{True}  & \texttt{False} \\
12 & 3+  & \texttt{True}  & \texttt{False} & \texttt{False} & \texttt{True}  \\
13 & 3+  & \texttt{True}  & \texttt{False} & \texttt{False} & \texttt{False} \\
14 & 3+  & \texttt{False} & \texttt{True}  & \texttt{True}  & \texttt{True}  \\
15 & 3+  & \texttt{False} & \texttt{True}  & \texttt{True}  & \texttt{False} \\
16 & 3+  & \texttt{False} & \texttt{True}  & \texttt{False} & \texttt{True}  \\
17 & 3+  & \texttt{False} & \texttt{True}  & \texttt{False} & \texttt{False} \\
18 & 3+  & \texttt{False} & \texttt{False} & \texttt{True}  & \texttt{True}  \\
19 & 3+  & \texttt{False} & \texttt{False} & \texttt{True}  & \texttt{False} \\
20 & 3+  & \texttt{False} & \texttt{False} & \texttt{False} & \texttt{True}  \\
21 & 3+  & \texttt{False} & \texttt{False} & \texttt{False} & \texttt{False} \\
\bottomrule
\end{tabular}
\end{table}

\section{Compute Information}
\label{app:compute}
We utilized a hardware configuration consisting of 8 NVIDIA A100-SXM4-40GB GPUs and 128 CPUs to collect data from Qwen-2-72B and Llama-3-70B. Using this setup, generating a single Qwen-2-72B player took an average of 7.5 minutes, as same as Llama-3-70B player. In total, we generated 4177 Qwen-2-72B players for our experiments, which took approximately 21.75 days to complete. To generate one LLM player using Llama-3-8B, we used a single GPU for 2 minutes. 
We have used API calls to the Google Cloud platform to generate players with Bison-chat and Gemini-1.5. Each LLM player of this setup costs around 0.5\$. 
We utilized one NVIDIA GeForce GTX 1080 GPU with 8GB of memory to train the prediction models. Training the LSTM model with 4096 LLM players took approximately 16 minutes.